\documentclass[lettersize,journal]{IEEEtran}
\usepackage{amsmath,amsfonts}
\usepackage{algorithmic}
\usepackage{algorithm}
\usepackage{array}
\usepackage[caption=false,font=normalsize,labelfont=sf,textfont=sf]{subfig}
\usepackage{textcomp}
\usepackage{stfloats}
\usepackage{url}
\usepackage{verbatim}
\usepackage{graphicx}
\usepackage{cite}

\usepackage{hyperref}       
\usepackage{url}            
\usepackage{booktabs}       
\usepackage{amsfonts}       
\usepackage{nicefrac}       
\usepackage{microtype}      
\usepackage{xcolor}         

\usepackage{multirow, makecell}
\usepackage{bbding}
\usepackage{pifont}
\usepackage{wasysym}
\usepackage{caption}
\usepackage{subcaption}
\usepackage{wrapfig}
\usepackage{enumitem}
\usepackage{amssymb}
\usepackage{cleveref}
\usepackage[misc]{ifsym}

\newcommand{\ie}{\textit{i}.\textit{e}.}
\newcommand{\eg}{\textit{e}.\textit{g}.}

\newcommand{\vs}{\textit{vs}.\ }

\begin{document}

\title{Pixel-Perfect Visual Geometry Estimation}

\author{\textbf{Gangwei Xu} \quad \textbf{Haotong Lin} \quad \textbf{Hongcheng Luo} \quad \textbf{Haiyang Sun} \quad \textbf{Bing Wang} \quad \\ \textbf{Guang Chen} \quad \textbf{Sida Peng} \quad \textbf{Hangjun Ye\dag} \quad  \textbf{Xin Yang\dag}

\thanks{Gangwei Xu and Xin Yang are with the School of Electronic Information and Communications,
Huazhong University of Science and Technology, Wuhan 430074, China.}

\thanks{Haotong Lin and Sida Peng are with the College of Computer Science and Technology, 
Zhejiang University, Hangzhou 310027, China.}

\thanks{Hongcheng Luo, Haiyang Sun, Bing Wang, Guang Chen, and Hangjun Ye are with Xiaomi EV, Beijing, 100081, China.}

\thanks{Corresponding author\dag: Xin Yang and Hangjun Ye.}}



\maketitle

\begin{abstract}
Recovering clean and accurate geometry from images is essential for robotics and augmented reality. However, existing geometry foundation models still suffer severely from flying pixels and the loss of fine details. In this paper, we present pixel-perfect visual geometry models that can predict high-quality, flying-pixel-free point clouds by leveraging generative modeling in the pixel space. We first introduce Pixel-Perfect Depth (PPD), a monocular depth foundation model built upon pixel-space diffusion transformers (DiT). To address the high computational complexity associated with pixel-space diffusion, we propose two key designs: 1) Semantics-Prompted DiT, which incorporates semantic representations from vision foundation models to prompt the diffusion process, preserving global semantics while enhancing fine-grained visual details; and 2) Cascade DiT architecture that progressively increases the number of image tokens, improving both efficiency and accuracy. To further extend PPD to video (PPVD), we introduce a new Semantics-Consistent DiT, which extracts temporally consistent semantics from a multi-view geometry foundation model. We then perform reference-guided token propagation within the DiT to maintain temporal coherence with minimal computational and memory overhead. Our models achieve the best performance among all generative monocular and video depth estimation models and produce significantly cleaner point clouds than all other models. Code is available at \url{https://github.com/gangweix/pixel-perfect-depth}.

\end{abstract}

\begin{figure*}
    \centering
    \includegraphics[width=1\linewidth]{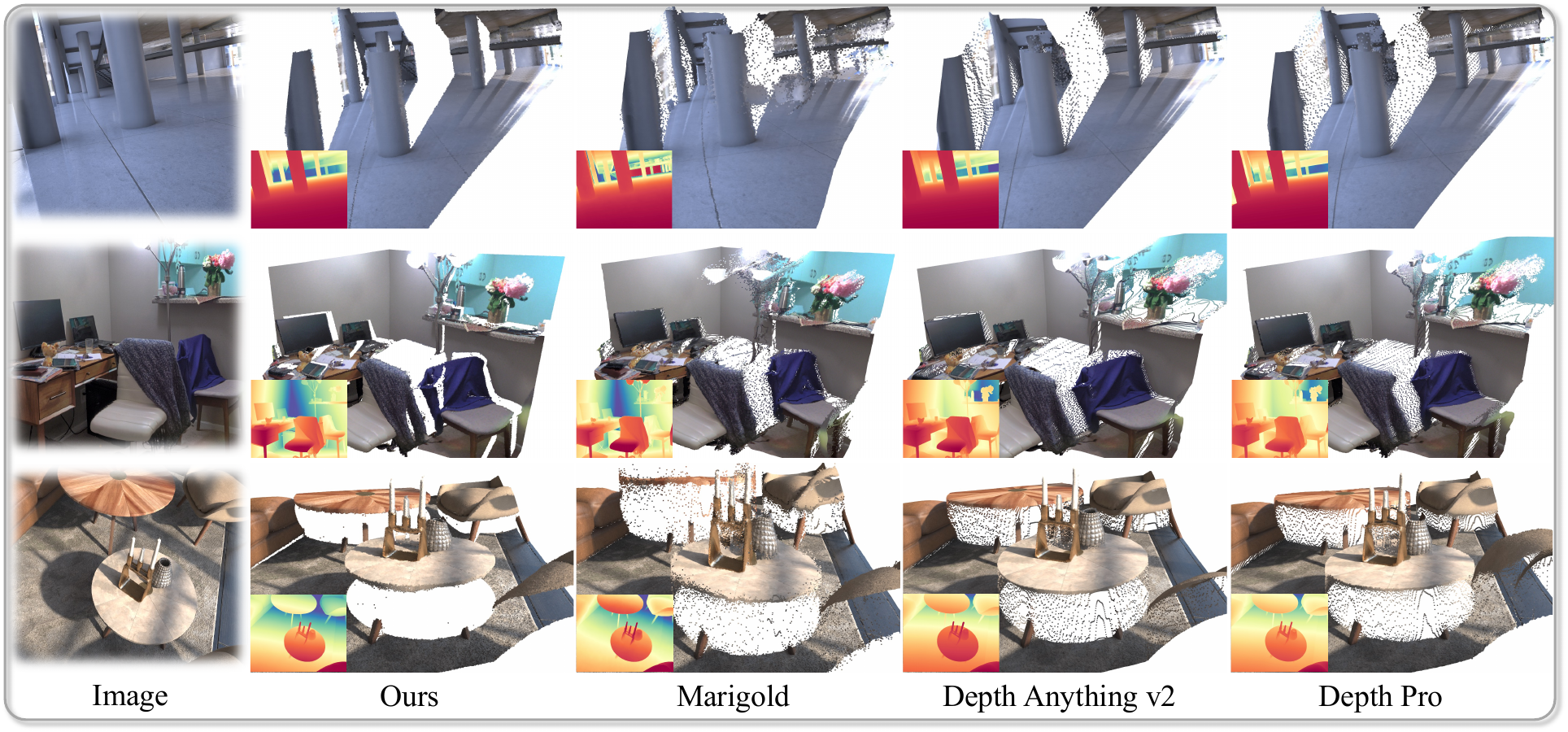}
    \caption{
    \textbf{Visual comparison with existing depth foundation models.} Discriminative models such as Depth Anything v2 and generative models such as Marigold, due to their inherent modeling paradigms or architectural limitations, produce substantial \textit{flying pixels}. In contrast, our model estimates depth maps that produce high-quality, flying-pixel-free point clouds without any additional refinement or post-processing.
    }
    \label{fig:teaser}
\end{figure*}

\section{Introduction}
\begin{figure*}
    \centering
    \includegraphics[width=1\linewidth]{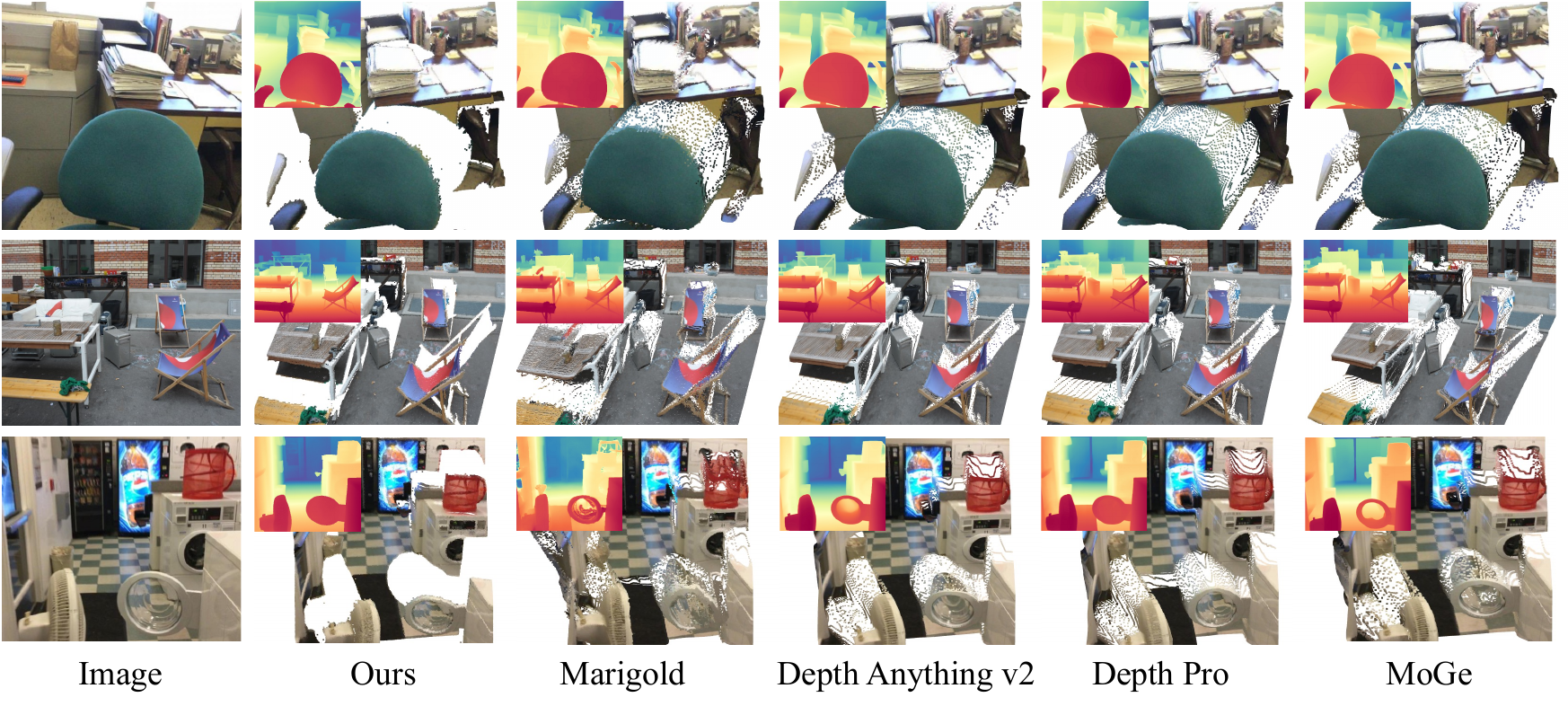}
    \caption{\textbf{Qualitative point cloud results of monocular depth estimation.}  Our PPD produces significantly fewer \textit{flying pixels} compared to other monocular depth models~\cite{ke2024marigold,yang2024depthv2,depthpro}, with depth maps overlaid on the point clouds for visualization.
    }
    \label{fig:pc_comp}
\end{figure*}

Monocular visual geometry estimation is a fundamental task with a wide range of downstream applications, such as robotics, autonomous driving, and augmented reality. Due to its significance, a large number of depth estimation models~\cite{ke2024marigold,yang2024depth,yang2024depthv2,moge2,yin2023metric3d,vda,depthcrafter,da3} have emerged recently. These models achieve impressive results in most zero-shot scenarios or regions, but suffer from \textit{flying pixels} around object boundaries and fine details when converted into point clouds~\cite{liang2025parameter}, as shown in Figure~\ref{fig:teaser} and ~\ref{fig:pc_comp}, which limits their practical applications in tasks such as high-precision robotic manipulation~\cite{maddukuri2025sim}, autonomous navigation~\cite{li2025recogdrive}, and immersive AR/VR rendering~\cite{enerf,depthsplat}.

\begin{figure*}
    \centering
    \includegraphics[width=1\linewidth]{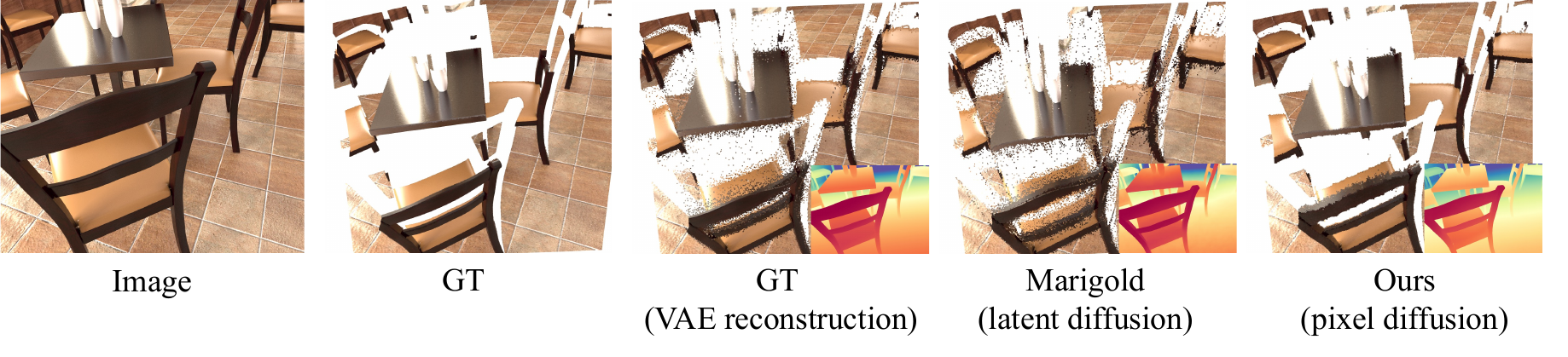}
    \caption{\textbf{Pixel diffusion \vs  latent diffusion.} GT(VAE reconstruction) denotes the ground truth depth map after VAE reconstruction. Existing generative models~\cite{ke2024marigold} use a VAE to compress inputs into the latent space, inevitably introducing \textit{flying pixels} at edges and details. In contrast, our model directly performs diffusion in pixel space, avoiding these issues. Depth maps are visualized on the point clouds.}
    \label{fig:vae}
\end{figure*}

Current geometry foundation models~\cite{ke2024marigold,yang2024depthv2,moge2, vda} suffer from the \textit{flying pixels} problem due to their inherent modeling paradigms and architectural limitations. For discriminative models, such as Depth Anything~\cite{yang2024depth,yang2024depthv2} and VGGT~\cite{vggt}, \textit{flying pixels} mainly arise from their tendency to predict intermediate (\textit{average}) depth values between the foreground and background at depth-discontinuous edges in order to minimize regression loss. In contrast, generative models such as Marigold~\cite{ke2024marigold} and DepthCrafter~\cite{depthcrafter} bypass direct regression by modeling pixel-wise depth distributions, enabling the recovery of sharper geometric edges and the more faithful reconstruction of fine structures. However, current generative depth models typically fine-tune Stable Diffusion~\cite{stablediff} for depth estimation, which requires a Variational Autoencoder (VAE) to compress depth maps into a latent space. This compression inevitably leads to the loss of edge sharpness and structural fidelity, resulting in a significant number of \textit{flying pixels}, as shown in Figure~\ref{fig:vae}. 

A trivial solution could be training a diffusion-based depth model in pixel space, bypassing the use of a VAE. However, we find this highly challenging, due to the increased complexity and instability of modeling both global semantic consistency and fine-grained visual details, leading to extremely low-quality depth predictions (Table~\ref{tab:ablation_mde} and Figure~\ref{fig:sp_dit}). Prior works have attempted to improve either the generative performance in high-resolution spaces or the training efficiency of diffusion-based models. For example, Simple Diffusion~\cite{hoogeboom2023simple} modifies the signal-to-noise ratio (SNR) to enhance high-resolution diffusion quality, while REPA~\cite{repa} improves training efficiency by aligning intermediate diffusion tokens with a pretrained vision encoder. However, these improvements remain limited and still fall short of enabling high-resolution pixel-space diffusion models to achieve performance comparable to state-of-the-art depth foundation models~\cite{yang2024depthv2,moge2}, as shown in Table~\ref{tab:ablation_mde}.
 

In this paper, we present \textbf{Pixel-Perfect Depth} (\textbf{PPD}), a framework for high-quality and flying-pixel-free monocular depth estimation using pixel-space diffusion transformers.
Recognizing that the major difficulty in high-resolution pixel-space diffusion lies in perceiving and modeling global image structures. To address this challenge, we propose the \textbf{Semantics-Prompted Diffusion Transformers} (\textbf{SP-DiT}) that incorporate high-level semantic representations into the diffusion process to enhance the model's ability to preserve global structures and semantic coherence. Equipped with SP-DiT, our model can more effectively preserve global semantics while generating fine-grained visual details in high-resolution pixel space. As shown in Table~\ref{tab:ablation_mde} and Figure~\ref{fig:sp_dit}, SP-DiT significantly improves overall performance, with up to a 78\% gain on the NYUv2~\cite{nyuv2} AbsRel metric.

Furthermore, we introduce the \textbf{Cascade DiT} architecture (\textbf{Cas-DiT}), an efficient architecture for diffusion transformers. We find that in diffusion transformers, the early blocks are primarily responsible for capturing and generating global or low-frequency structures, while the later blocks focus on generating high-frequency details. Based on this insight, Cas-DiT adopts a progressive patch size strategy: larger patch size is used in the early DiT blocks to reduce the number of tokens and facilitate global image structure modeling; in the later DiT blocks, we increase the number of tokens, which is equivalent to using a smaller patch size, allowing the model to focus on the generation of fine-grained spatial details. This coarse-to-fine cascaded architecture not only significantly reduces computational costs but also improves efficiency.

A preliminary version of this work was published at NeurIPS 2025. However, the conference version suffers from a notable limitation: it lacks temporal consistency when applied to long videos, resulting in flickering depth predictions. In this paper, we extend \textbf{PPD} to arbitrarily long video sequences, which we term \textbf{Pixel-Perfect Video Depth} (\textbf{PPVD}). Previous video depth estimation models~\cite{depthcrafter,rollingdepth,vda} suffer from two limitations: first, they consider only temporal propagation and do not perform joint spatiotemporal (global) propagation; second, they ignore camera motion, which causes temporal propagation to transfer incorrect semantics and thus hinders performance.


To achieve high temporal consistency, strong spatial accuracy, and well-preserved details, we propose a novel \textbf{Semantics-Consistent DiT} (\textbf{SC-DiT}). SC-DiT integrates view-consistent semantics extracted from a multi-view geometry foundation model~\cite{vggt,pi3,da3,mapanything} into the DiT. These semantics provide strong 3D reconstruction consistency while implicitly encoding camera motion.  Moreover, instead of relying on direct global propagation, \ie,  computationally expensive full attention over all frames ($T \times H \times W$), SC-DiT introduces a Reference-Guided Token Propagation (RGTP) strategy, enabling temporal consistency while using only single-frame self-attention. Specifically, RGTP first assigns sparse (compressed) reference-frame tokens to all video frames, and then performs self-attention only on single-frame tokens. Through these sparse reference tokens, the scene’s scale and shift information can be propagated throughout the entire video sequence. Finally, PPVD outperforms the previous best method, Video Depth Anything~\cite{vda}, by 38.7\% and 58.4\% on the NYUv2 and ScanNet benchmarks, respectively.

We highlight the main contributions of this paper below:
\begin{itemize}[leftmargin=*, itemsep=0.5em]

\item We present Pixel-Perfect Visual Geometry estimation models, including \textbf{PPD} for monocular depth estimation and \textbf{PPVD} for video depth estimation, both capable of producing flying-pixel-free point clouds from the estimated depth maps.

    
\item We propose Semantics-Prompted DiT for PPD and Semantics-Consistent DiT for PPVD. The former substantially improves accuracy and enhances fine details, while the latter not only boosts accuracy but also strengthens temporal consistency. In addition, a Cascaded DiT architecture is employed to further enhance their efficiency.

\item We introduce a Reference-Guided Token Propagation strategy, enabling single-view self-attention to propagate global spatiotemporal information, thereby maintaining temporal consistency while minimizing computational overhead.

\item Our PPD and PPVD set new state-of-the-art results among generative monocular and video depth estimation models. Moreover, to effectively assess \textit{flying pixels} at object edges, we introduce an edge-aware point cloud evaluation metric, on which our models achieve the best performance.

\end{itemize}

\section{Related Work}
\label{related_work}

\subsection{Monocular Depth Estimation}
Depth estimation can be broadly categorized into monocular~\cite{yang2024depthv2,wang2025moge}, stereo~\cite{dust3r, fast3r, igev,wang2024selective, xu2023accurate, igevpp, guo2023openstereo, cheng2025monster, cheng2024adaptive}, and sparse depth completion~\cite{promptda,marigold-dc} methods.
Early monocular depth estimation methods relied primarily on manually designed features~\cite{saxena2008make3d,hoiem2007recovering}. 
The advent of neural networks revolutionized the field, though initial approaches~\cite{eigen2014depth,eigen2015predicting} 
struggled with cross-dataset generalization.
To address this limitation, scale-invariant and relative loss~\cite{midas} are introduced, enabling multi-dataset~\cite{Li2018:CVPR,blendedmvs,diml,hrwsi,tartanair,irs,wsvd,redweb,hypersim,liang2025sood++} training.
Recent methods focus on improving the 
generalization ability~\cite{yang2024depthv2,wang2025jasmine,wang2025editor}, 
depth consistency~\cite{depthanyvideo, vda, depthcrafter,flashdepth},  
and metric scale~\cite{bhat2023zoedepth,li2024patchfusion,li2024patchrefiner,yin2023metric3d,guizilini2023towards,leres,hu2024metric3dv2,piccinelli2024unidepth, promptda} 
of depth estimation. These methods converge towards using transformer-based architectures~\cite{dpt}. Among them, MoGe~\cite{wang2025moge} has achieved high accuracy and strong generalization. However, it also suffers from \textit{flying pixels} and the loss of fine details. Depth Pro~\cite{depthpro} improves detail recovery by increasing the input image resolution, yet its generalization remains limited when applied to diverse real-world scenes. Several recent methods~\cite{ji2023ddp,duan2024diffusiondepth,saxena2023monocular,saxena2023surprising,saxena2023zero,zhao2023unleashing} have attempted to use diffusion models for metric depth estimation. But, they struggle to generalize to real-world scenes and lose fine-grained details.

Most recently, ~\cite{ke2024marigold} brought the new insight to the field by fine-tuning pretrained Stable Diffusion~\cite{stablediff} for depth estimation, which demonstrated impressive zero-shot capabilities for relative depth. The following works~\cite{he2024lotus,depthfm,song2025depthmaster,zhang2024betterdepth,bai2024fiffdepth} attempt to improve its performance and inference speed.
However, they are all based on the latent diffusion model~\cite{stablediff}, which is trained in the latent space and requires a VAE to compress the depth map into a latent space. Moreover, the compression inherent in VAE inevitably leads to a large number of \textit{flying pixels}. We focus on a pixel-space diffusion model that is trained directly in the pixel space without requiring any VAE. As a result, our model is able to produce high-quality and flying-pixel-free point clouds from the estimated depth maps.

\subsection{Video Depth Estimation}
Although monocular depth foundation models~\cite{yang2024depthv2,moge2} exhibit strong generalization ability, they commonly suffer from temporal flickering. The goal of video depth estimation is to achieve temporal stability while preserving spatial accuracy. Early video depth methods~\cite{kopf2021robust,luo2020consistent} relied on test-time optimization, which are impractical for real-world deployment. Subsequent learning-based work, such as NVDS~\cite{nvds}, employs a stabilization network to directly predict temporally consistent depth from videos, improving inference efficiency. However, its generalization ability is constrained by the limited diversity of the training data and the model capacity. 

Recently, several works, such as~\cite{depthcrafter,depthanyvideo,chorondepth}, have leveraged pretrained video diffusion models~\cite{svd} for video depth estimation, achieving strong generalization to real-world scenes. However, they often consider only local temporal propagation and fail to perform joint spatiotemporal (global) propagation. This limitation can lead to the propagation of incorrect semantics, consequently resulting in poor spatial accuracy. Instead of using video diffusion models, RollingDepth~\cite{rollingdepth} fine-tunes an image diffusion model and then applies an optimization-based co-alignment procedure for video depth. Moreover, these generative depth estimation models all rely on a VAE, which inevitably introduces \textit{flying pixels}. To improve inference efficiency, Video Depth Anything~\cite{vda} is built on top of Depth Anything~\cite{yang2024depthv2} and introduces a lightweight spatial–temporal head to enforce temporal consistency. However, its emphasis on temporal smoothness comes at the cost of spatial accuracy. In contrast, our PPVD elegantly converts 3D geometry consistency into temporal consistency, enabling temporal stability while preserving high spatial accuracy.

\subsection{Diffusion Generative Models}
Diffusion generative models~\cite{ddpm,ddim,dit,repa, yao2024fasterdit,yao2025reconstruction,zhu2025dig} have demonstrated impressive results in image and video generation. Early approaches~\cite{ddpm,ho2022classifier,ho2022cascaded} such as DDPM~\cite{ddpm} operate directly in the pixel space, enabling high-fidelity generation but incurring significant computational costs, especially at high resolutions. 
To address this limitation, Latent Diffusion Models perform the diffusion process in a lower-dimensional latent space obtained via a VAE, as popularized by Stable Diffusion~\cite{stablediff}. This design significantly improves training and inference efficiency and has been widely adopted in recent works~\cite{sd3,yao2025reconstruction,repa, zhu2024dig,flux2024, podell2023sdxl,yang2024cogvideox,wan2025wan}.

Diffusion models for depth estimation typically share a similar design. For example, Marigold~\cite{ke2024marigold} and its follow-up works~\cite{he2024lotus,depthfm,depthcrafter} fine-tune pretrained Stable Diffusion~\cite{stablediff} or Stable Video Diffusion~\cite{svd} models for monocular or video depth estimation, benefiting from fast convergence and strong priors learned from large-scale datasets. However, the VAE compression they rely on inevitably introduces \textit{flying pixels} in the resulting point clouds. In contrast, pixel-space diffusion avoids such artifacts but remains computationally intensive and slow to converge at high resolutions. To address these issues, we propose Semantics-Prompted DiT and Semantics-Consistent DiT, which enable depth estimation that is both flying-pixel-free and temporally consistent.

\section{Method}
\label{method}

\begin{figure*}
    \centering
    \includegraphics[width=0.8\linewidth]{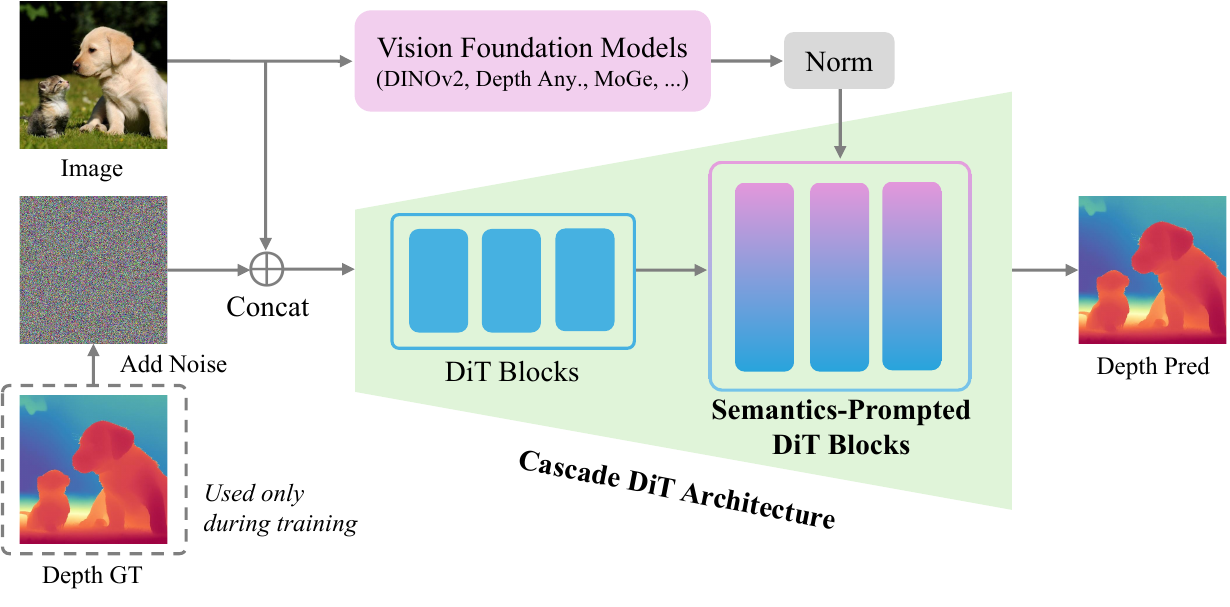}
    \caption{\textbf{Overview of Pixel-Perfect Depth.} Given an input image concatenated with noise, we feed it into the proposed Cascade DiT. The image is also processed by a pretrained encoder from Vision Foundation Models to extract high-level semantics, forming our Semantics-Prompted DiT. We perform diffusion generation directly in pixel space without using any VAE.
    }
    \label{fig:network}
\end{figure*}

\subsection{Pixel-Perfect Depth \& Pixel-Perfect Video Depth}
\label{ppd_ppvd}
Given a single image or a video sequence, our goal is to estimate pixel-perfect monocular or video depth that produces flying-pixel-free point clouds. Existing depth foundation models~\cite{ke2024marigold,fu2025geowizard,he2024lotus,yang2024depthv2,depthpro} universally suffer from \textit{flying pixels}, stemming from their inherent modeling paradigms or architectural limitations. For example, discriminative models, although achieving significantly higher accuracy than generative ones, tend to smooth object edges and blur fine details due to their mean-prediction bias, which in turn leads to \textit{flying pixels}. Generative models, in principle, can better capture the multi-modal depth distributions around object boundaries and fine details. However, current generative models typically fine-tune latent diffusion models~\cite{stablediff,svd} for depth estimation, requiring the depth map to be compressed into a latent space via a VAE, which inevitably introduces \textit{flying pixels}.

To unleash the potential of generative models for depth estimation, we propose \textbf{Pixel-Perfect Depth} that performs diffusion directly in the pixel space instead of the latent space. It allows us to directly model the pixel-wise distribution of depth, such as the discontinuities at object edges.
However, training a generative diffusion model directly in the high-resolution pixel space (\eg, 1024$\times$768) is computationally demanding and hard to optimize. To overcome these challenges, we introduce Semantics-Prompted DiT (SP-DiT), detailed in Section~\ref{sec:sp_dit}.

While Semantics-Prompted DiT enables our pixel-space diffusion model for monocular depth estimation to train effectively and achieve state-of-the-art performance, its direct application to video still results in noticeable temporal flickering. To enable our model to perform effectively on video, we propose Semantics-Consistent DiT, whose core idea is to transform 3D geometry reconstruction consistency into temporal consistency. To enforce temporal consistency in DiT efficiently, we introduce a reference-guided token propagation strategy that performs single-view self-attention to propagate global spatiotemporal information at minimal computational cost, detailed in Section~\ref{sec:sc_dit}.

\begin{figure*}
    \centering
    \includegraphics[width=0.8\linewidth]{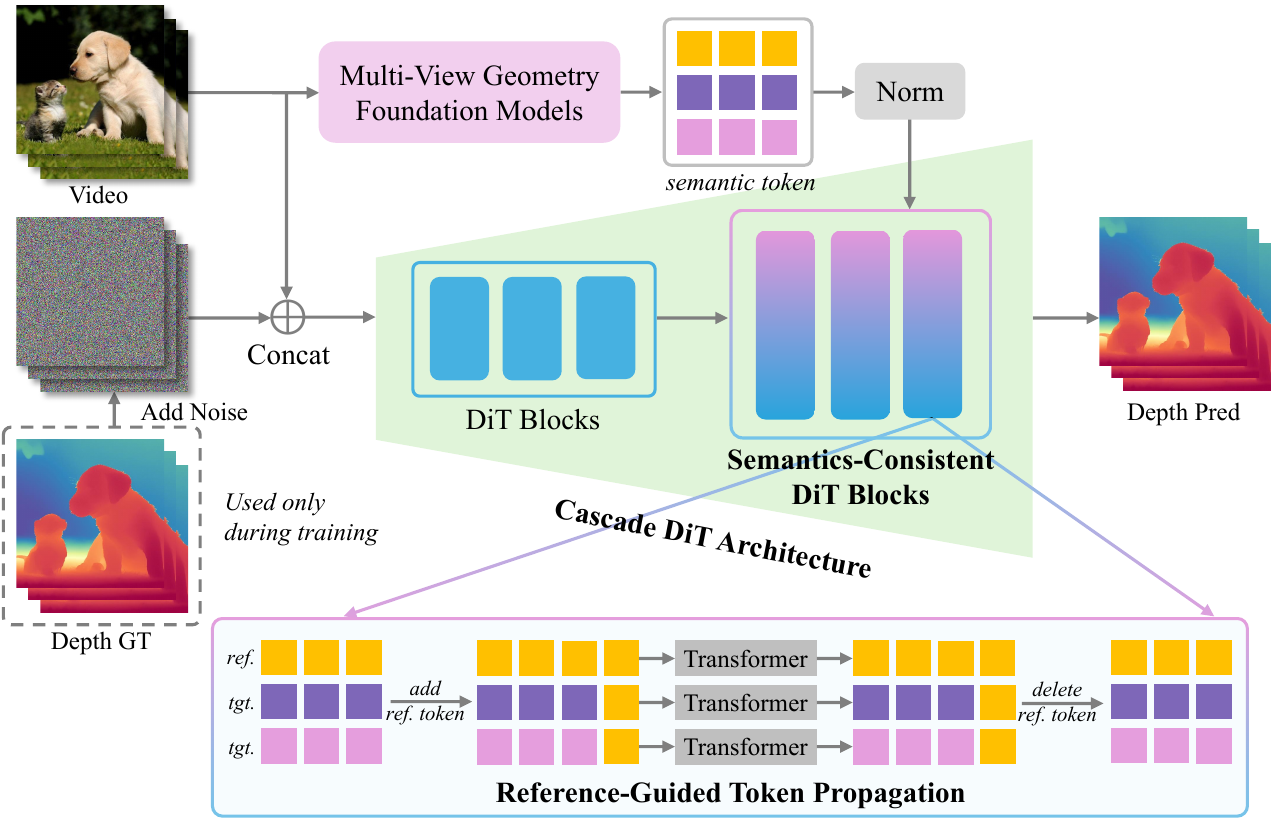}
    \caption{\textbf{Overview of Pixel-Perfect Video Depth.} Given a sequence of video frames concatenated with noise, we feed it into the proposed Cascade DiT. The video is also processed by a multi-view geometry-based model to capture spatiotemporally consistent semantics, forming our Semantics-Consistent DiT. In the subsequent DiT, to ensure temporal coherence within the single-view transformer, we introduce a reference-guided token propagation strategy, where sparse reference tokens propagate scale and shift information across frames.
    }
    \label{fig:ppvd_network}
\end{figure*}

\subsection{Generative Formulation}

We adopt Flow Matching~\cite{lipman2022flow,liu2022flow,albergo2022building} as the generative core of our depth estimation framework. Flow Matching learns a continuous transformation from Gaussian noise to a data sample via a first-order Ordinary Differential Equation (ODE). In our case, we model the transformation from Gaussian noise to a depth sample. Specifically, given a clean depth sample $\mathbf{x}_0 \sim \mathcal{D}$ and Gaussian noise $\mathbf{x}_1 \sim \mathcal{N}(0, 1)$, we define an interpolated sample at continuous time $t \in [0, 1]$ as:
\begin{equation}
\mathbf{x}_t = t \cdot \mathbf{x}_1 + (1 - t) \cdot \mathbf{x}_0.
\end{equation}
This defines a velocity field:
\begin{equation}
\mathbf{v}_t = \frac{d \mathbf{x}_t}{dt} = \mathbf{x}_1 - \mathbf{x}_0,
\end{equation}
which describes the direction from clean data to noise. Our model $\mathbf{v}_\theta(\mathbf{x}_t, t, \mathbf{c})$ is trained to predict the velocity field, based on the current noisy sample $\mathbf{x}_t$, the time step $t$, and the input image $\mathbf{c}$. The training objective is the mean squared error (MSE) between the predicted and true velocity:
\begin{equation}
\mathcal{L}_{\text{velocity}(\theta)} = \mathbb{E}_{\mathbf{x}_0, \mathbf{x}_1, t} \left[ \left\| \mathbf{v}_{\theta}(\mathbf{x}_t, t, \mathbf{c}) - \mathbf{v}_t \right\|^2 \right].
\label{eq:loss}
\end{equation}

At inference, we start from noise $\mathbf{x}_1$ and solve the ODE by discretizing the time interval $[0, 1]$ into steps ${t_i}$, iteratively updating the depth sample as follows:
\begin{equation}
    \mathbf{x}_{t_{i-1}} = \mathbf{x}_{t_i} + \mathbf{v}_\theta(\mathbf{x}_{t_i}, t_i, \mathbf{c})(t_{i-1} - t_i),
\end{equation}
where $t_i$ decreases from 1 to 0, gradually transforming the initial noise $\mathbf{x}_1$ into the depth sample $\mathbf{x}_0$.

\begin{figure*}
    \centering
    \includegraphics[width=0.85\linewidth]{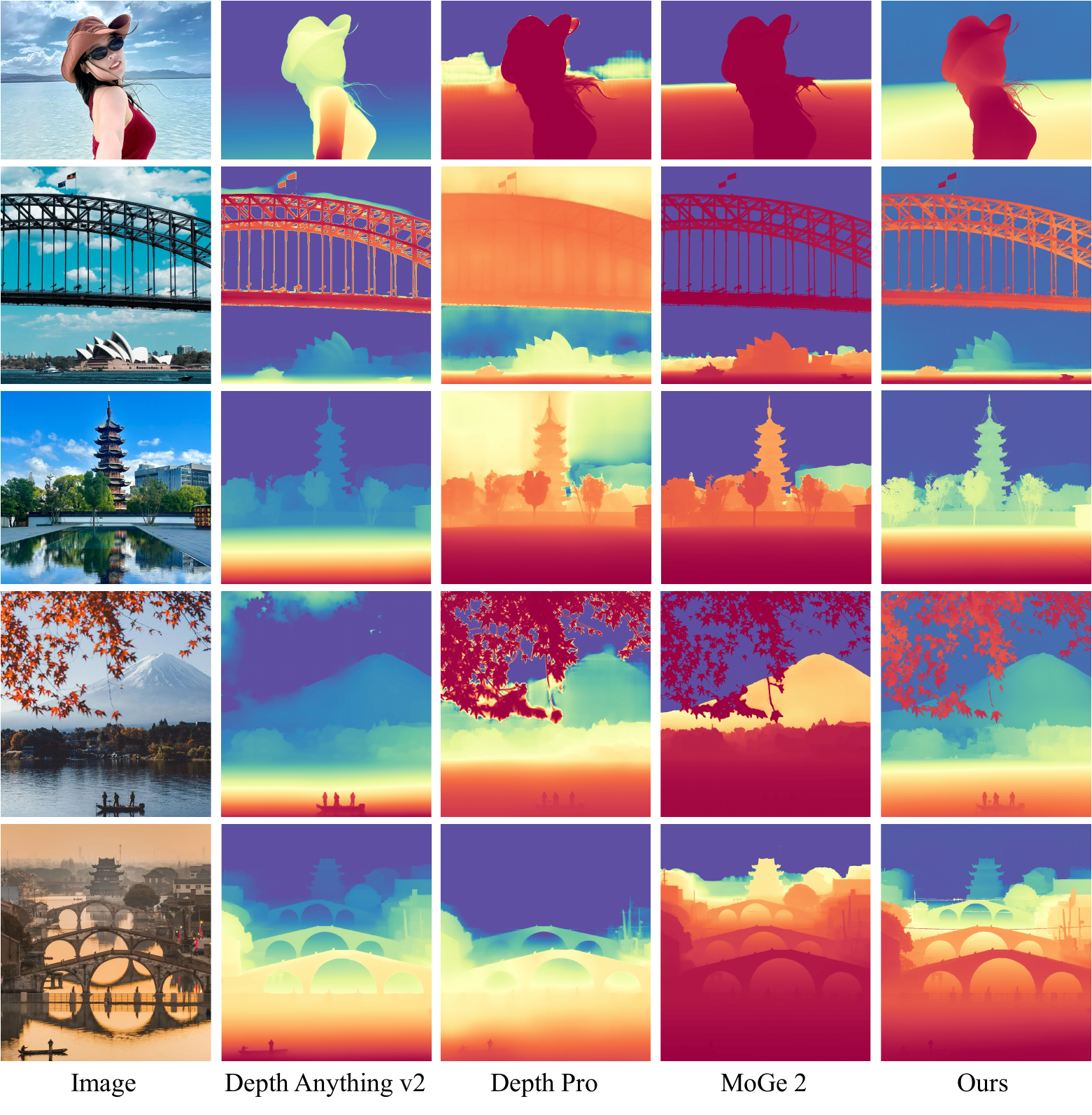}
    \caption{\textbf{Comparison with existing depth foundation models.} Our PPD preserves more fine-grained details than Depth Anything v2~\cite{yang2024depthv2} and MoGe 2~\cite{moge2}, while demonstrating significantly higher robustness compared to Depth Pro~\cite{depthpro}.}
    \label{fig:demo}
\end{figure*}

\subsection{Semantics-Prompted Diffusion Transformers}
\label{sec:sp_dit}
Our Semantics-Prompted DiT builds on DiT~\cite{dit} for its simplicity, scalability, and strong performance in generative modeling. Unlike previous depth estimation models such as Depth Anything v2~\cite{yang2024depthv2} and Marigold~\cite{ke2024marigold}, our architecture is purely transformer-based, without any convolutional layers. By integrating high-level semantic representations, SP-DiT enables our model to preserve spatial semantic consistency while enhancing fine-grained visual details, without sacrificing the simplicity and scalability of DiT.

Specifically, given the interpolated noise sample $\mathbf{x}_t$ and the corresponding image $\mathbf{c}$, we first concatenate them into a single input: $\mathbf{a}_t=\mathbf{x}_t \oplus \mathbf{c}$, where the image $\mathbf{c}$ serves as a condition. Then, we directly feed $\mathbf{a}_t$ into the DiT. The first layer of DiT is a patchify operation, which converts the spatial input $\mathbf{a}_t$ into a 1D sequence of $T$ tokens (patches), each with a dimension of $D$, by linearly embedding each patch of size $p \times p$ from the input $\mathbf{a}_t$.
Subsequently, the input tokens are processed by a sequence of Transformer blocks, called DiT blocks. After the final DiT block, each token is linearly projected into a $p\times p$ tensor, which is then reshaped back to the original spatial resolution to obtain the predicted velocity $\mathbf{v}_t$ (\ie, $\mathbf{x}_1-\mathbf{x}_0$), with a channel dimension of 1.

Unfortunately, performing diffusion directly in the pixel space leads to poor convergence and highly inaccurate depth predictions. As shown in Figure~\ref{fig:sp_dit}, the model struggles to model both global image structure and fine-grained details. To address this, we extract high-level semantic representations $\mathbf{e}$ as guidance from the input image $\mathbf{c}$ using a vision foundation model $f$, as follows:
\begin{equation}
\mathbf{e}=f(\mathbf{c})\in\mathbb{R}^{T' \times D'},
\end{equation}
where $T'$ and $D'$ are the number of tokens and the embedding dimension of $f(\mathbf{c})$, respectively. These high-level semantic representations are then incorporated into our DiT model, enabling it to more effectively preserve spatial semantic consistency while enhancing fine-grained visual details. However, we found that the magnitude of the obtained semantics $\mathbf{e}$ differs significantly from the magnitude of the tokens in our DiT model, which affects both the stability of the model's training and its performance. To address this, we normalize the semantic representation $\mathbf{e}$ along the feature dimension using L2 norm, as follows:
\begin{equation}
\hat{\mathbf{e}} = \frac{\mathbf{e}}{\|\mathbf{e}\|_2}.
\end{equation}

Subsequently, the normalized semantic representation is integrated into the tokens $\mathbf{z}$ of our DiT model via a multilayer perceptron (MLP) layer $h_\phi$, 
\begin{equation}
\mathbf{z'} = h_\phi(\mathbf{z} \oplus \mathcal{B}(\hat{\mathbf{e}})),
\end{equation}
where $\mathcal{B}(\cdot)$ denotes the bilinear interpolation operator, which aligns the spatial resolution of the semantic representation $\hat{\mathbf{e}}$ with that of the DiT tokens. The resulting $\mathbf{z}'$ denotes the DiT tokens enhanced with semantics. After the fusion, the subsequent DiT blocks are prompted by semantics to effectively preserve spatial semantic consistency while enhancing fine-grained details in the high-resolution pixel space. We refer to these subsequent DiT blocks as Semantics-Prompted DiT. 

In this work, we experiment with various pretrained vision foundation models, including DINOv2~\cite{dinov2}, MAE~\cite{he2022masked}, Depth Anything v2~\cite{yang2024depthv2}, and MoGe 2~\cite{moge2}. All of them significantly boost performance and facilitate more stable and efficient training, as shown in Table~\ref{tab:ablation_vfm}. Note that we only utilize the encoder of each vision foundation model, \eg, a 24-layer Vision Transformer (ViT-L/14) for Depth Anything v2~\cite{yang2024depthv2}.

\subsection{Semantics-Consistent Diffusion Transformers}
\label{sec:sc_dit}
Although SP-DiT substantially enhances monocular depth accuracy, inconsistencies in semantics across video frames persist, leading to noticeable flickering in video depth. Instead of constraining semantics using optical flow or pose priors, we observe that current multi-view geometry foundation models~\cite{vggt,da3} achieve remarkable reconstruction consistency. Motivated by this, our goal is to transform multi-view reconstruction consistency into temporal consistency for video. 

To this end, we first employ a pretrained multi-view geometry foundation model to extract semantics from video frames that are consistent across viewpoints, while also implicitly encoding camera poses. In contrast, prior video depth estimation models such as DepthCrafter~\cite{depthcrafter} and Video Depth Anything~\cite{vda} do not incorporate camera poses, even though pose information is crucial for achieving temporally consistent video depth. Subsequently, we incorporate these consistent semantics into the DiT through a normalization module and an MLP layer, as described in Section~\ref{sec:sp_dit}. However, in the DiT, it is challenging to maintain consistency among tokens from different video frames. A straightforward approach would be to perform transformer over all spatiotemporal tokens ($T\times H \times W$), but this is computationally and memory intensive, especially for diffusion in pixel space. 

To efficiently perform spatiotemporal transformer operations, we propose a new reference-guided token propagation strategy. As illustrated in Figure~\ref{fig:ppvd_network}, before each Transformer layer, we downsample the tokens of the reference frame by a factor of 4 and concatenate them with all input frames. In this way, the reference frame serves as an information conduit that is propagated to all frames, allowing DiT to operate on each frame individually while preserving temporal consistency and minimizing computational and memory cost.

\subsection{Cascade DiT Architecture}
While SP-DiT significantly improves the spatial accuracy of monocular depth estimation and SC-DiT further enhances both spatial accuracy and temporal consistency, performing diffusion directly in pixel space remains computationally expensive. To address this issue, we propose a novel Cascaded DiT architecture to reduce the computational burden of the diffusion model. We observe that in DiT architectures, the early blocks are primarily responsible for capturing global image structures and low-frequency information, while the later blocks focus on modeling fine-grained, high-frequency details.

To optimize the efficiency and effectiveness of this process, we adopt a large patch size in the early DiT blocks. This design significantly reduces the number of tokens that need to be processed, leading to lower computational cost. Additionally, it encourages the model to prioritize learning and modeling global image structures and low-frequency information, which also better aligns with the high-level semantic representations extracted from the input image. In the later DiT blocks, we increase the number of tokens, which is equivalent to using a smaller patch size. This allows the model to better focus on fine-grained spatial details. The resulting coarse-to-fine cascaded design mirrors the hierarchical nature of visual perception and improves both the efficiency and accuracy of depth estimation.

Specifically, for our diffusion model with a total of $N$ DiT blocks, the first $N/2$ blocks constitute the coarse stage with a larger patch size, while the remaining $N/2$ blocks (\ie, SP-DiT or SC-DiT) form the fine stage using a smaller patch size.

\subsection{Implementation Details}
In this section, we provide essential information about the model architecture details, depth normalization, and training details.

\textbf{Model architecture details.} In our implementation, we use a total of $N=24$ DiT blocks, each operating at a hidden dimension of $D=1024$. The first 12 blocks are standard DiT blocks with a patch size of 16, corresponding to $(H/16)\times(W/16)$ tokens for an input of size $H\times W$. After the 12th block, we employ an MLP layer to expand the hidden dimension by a factor of 4, followed by reshaping to obtain $(H/8)\times(W/8)$ tokens. The remaining 12 SP-DiT (or SC-DiT) blocks then further process these $(H/8)\times(W/8)$ tokens. Finally, we employ an MLP layer followed by a reshaping operation to transform the processed tokens into an $H \times W$ depth map. In contrast to prior depth estimation models, such as Depth Pro~\cite{depthpro} and Video Depth Anything~\cite{vda}, our model does not rely on any convolutional layers.

\textbf{Depth normalization.} The ground truth depth values are normalized to match the scale expected by the diffusion model. Before normalization, we convert the depth values into log scale to ensure a more balanced capacity allocation across both indoor and outdoor scenes. Specifically, we apply the transformation $\tilde{\mathbf{d}} = \log(\mathbf{d} + \epsilon)$, 
where \( \tilde{\mathbf{d}} \) denotes the transformed depth, \( \mathbf{d} \) is the original depth value, and \( \epsilon \) is a small positive constant (\eg, 1) to ensure numerical stability. We then normalize the log-scaled depth \( \tilde{\mathbf{d}} \) using:
\begin{equation}
\hat{\mathbf{d}} = \frac{\tilde{\mathbf{d}} - d_{\min}}{d_{\max} - d_{\min}} - 0.5,
\end{equation}
where \(d_{min}\) and \(d_{max}\) denote the lower and upper depth percentiles of each map, respectively. For video depth estimation, we convert depth to its disparity representation, which is more stable for distant regions in videos.

\begin{figure*}
    \centering
    \includegraphics[width=1\linewidth]{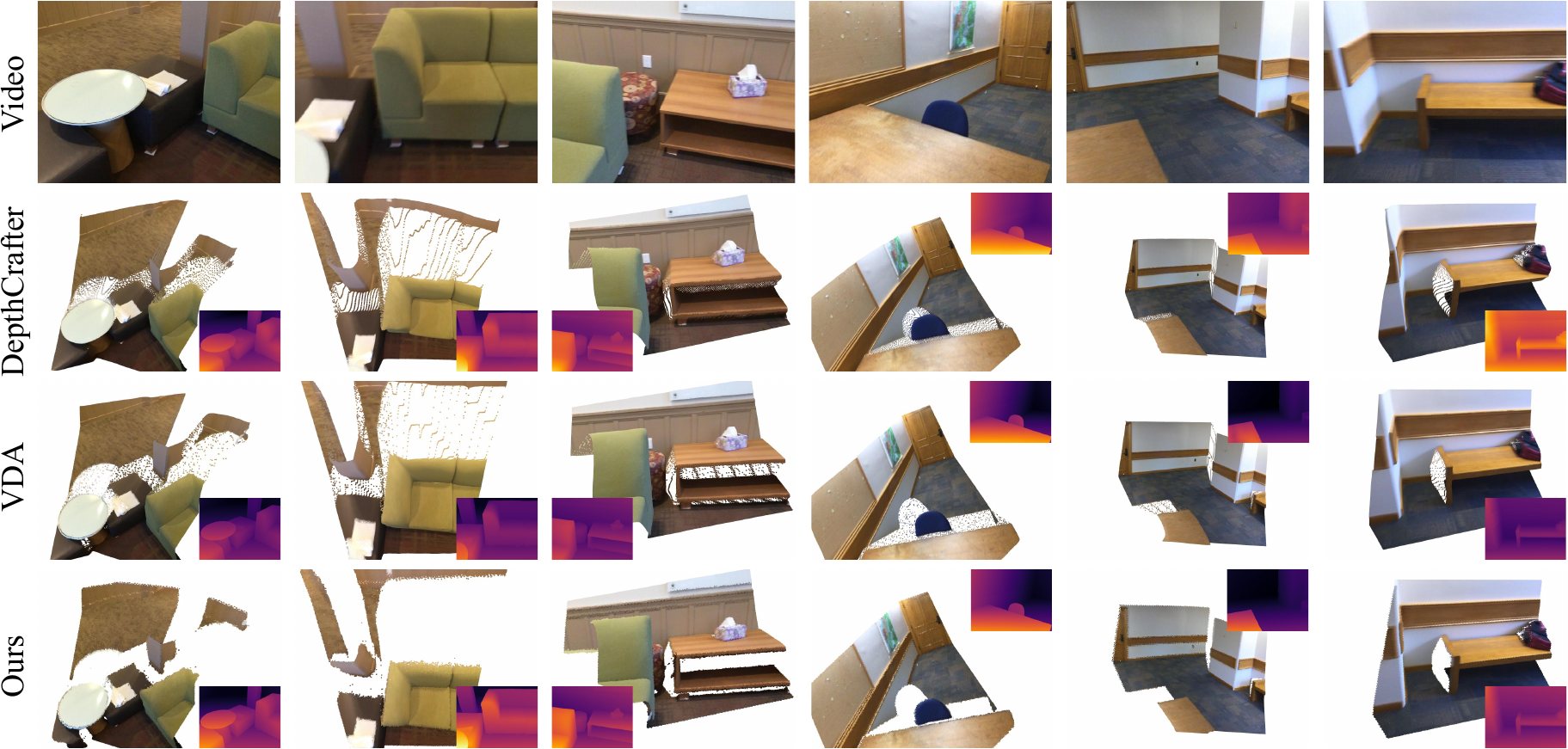}
    \caption{\textbf{Qualitative point cloud results of video depth estimation.}  Our PPVD produces significantly fewer \textit{flying pixels} compared to DepthCrafter~\cite{depthcrafter} and Video Depth Anything (VDA)~\cite{vda}, with depth maps overlaid on the point clouds.
    }
    \label{fig:pc_comp_video}
\end{figure*}

\begin{figure}
    \centering
    \includegraphics[width=1\linewidth]{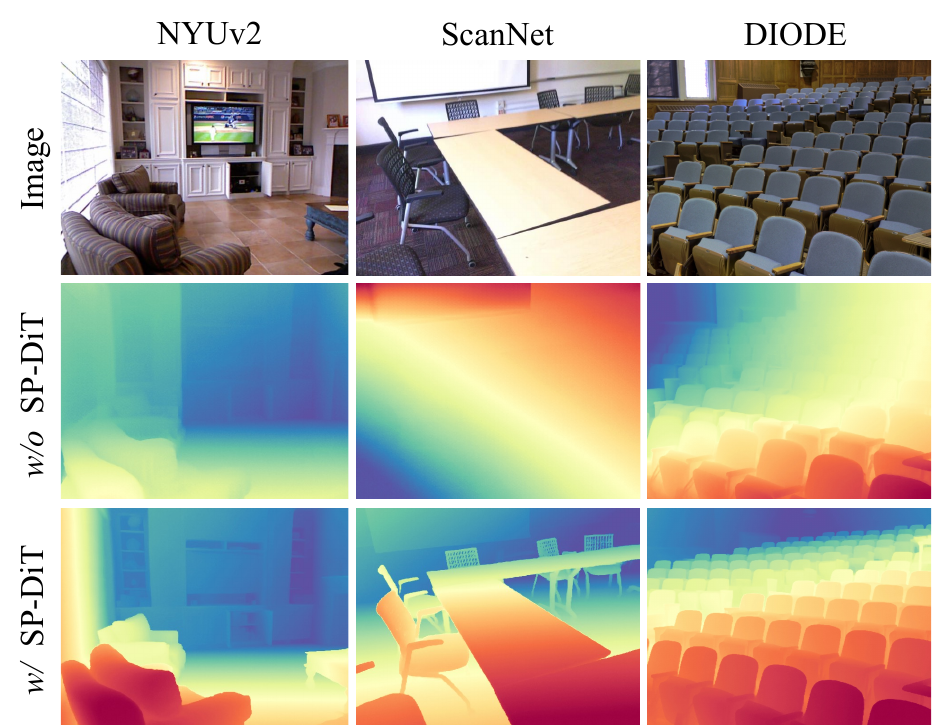}
    \caption{\textbf{Qualitative ablations for the proposed SP-DiT.} Without SP-DiT, the vanilla DiT model struggles with preserving global semantics and generating fine-grained visual details.}
    \label{fig:sp_dit}
\end{figure}

\textbf{Training details.} We introduce a progressive training strategy to stabilize optimization and improve training efficiency. For monocular depth estimation, we first train at a low resolution of $512 \times 512$ until convergence, and then fine-tune the model at a higher resolution of $1024 \times 768$. For video depth estimation, we begin by training on monocular images without the reference-guided token propagation strategy, and subsequently fine-tune the model on video sequences. The training losses are also designed progressively. During the pretraining stage, we use only the MSE loss between the predicted and ground-truth velocity, as shown in Equation~\ref{eq:loss}. In the fine-tuning stage, we further incorporate a gradient matching (GM) loss, adopted from~\cite{yang2024depthv2}. 

Specifically, for video depth estimation, we additionally propose a reference-aligned temporal gradient (RTG) loss, which complements our reference-guided token propagation strategy. This loss is computed as,
\begin{equation}
\mathcal{L}_{\mathrm{RTG}}\!=\!\frac{1}{R(T\!-\!R)} \sum_{j=R+1}^{T\!-\!R}\sum_{i=1}^{R}\| (\mathbf{d}_{j}^{pr} - \mathbf{d}_{i}^{pr})\! -\! (\mathbf{d}_{j}^{gt} - \mathbf{d}_{i}^{gt}) \|_1,
\end{equation}
where $T$ denotes the length of the input video clip, $R$ denotes the length of the reference frames, $\mathbf{d}^{pr}$ represents the depth prediction, and $\mathbf{d}^{gt}$ represents the ground-truth depth. In our experiments, we set $T=16$ and $R=3$. 

Finally, for monocular depth estimation, the total loss is defined as follows:
\begin{equation}
\mathcal{L}_{\mathrm{MDE}}=\mathcal{L}_{\mathrm{MSE}} + \alpha \mathcal{L}_{\mathrm{GM}}.
\end{equation}
For video depth estimation, the total loss is defined as follows:
\begin{equation}
\mathcal{L}_{\mathrm{VDE}}=\mathcal{L}_{\mathrm{MSE}} + \alpha \mathcal{L}_{\mathrm{GM}} + \beta \mathcal{L}_{\mathrm{RTG}},
\end{equation}
where $\alpha$ and $\beta$ are the weights used to balance temporal consistency and spatial accuracy. We train all models on 8 NVIDIA GPUs, using the AdamW optimizer with a constant learning rate of \(1\times 10^{-4}\).


\begin{table*}
    \caption{\textbf{Zero-shot monocular depth estimation.} Better: AbsRel $\downarrow$, $\delta_1$ $\uparrow$. \textbf{Bold} numbers are the best. Our PPD significantly outperforms all other generative depth models on five benchmarks. All metrics are presented in percentage terms.}
    \small
    \centering
    \begin{tabular}{lccccccccccc}
    \toprule
    \multirow{2}{*}{Type} & \multirow{2}{*}{Method} & \multicolumn{2}{c}{NYUv2} & \multicolumn{2}{c}{KITTI} & \multicolumn{2}{c}{ETH3D} & \multicolumn{2}{c}{ScanNet} & \multicolumn{2}{c}{DIODE} \\
    \cmidrule(lr){3-4}\cmidrule(lr){5-6}\cmidrule(lr){7-8}\cmidrule(lr){9-10}\cmidrule(lr){11-12}
    
      &  &AbsRel$\downarrow$ & $\delta_1$$\uparrow$ & AbsRel$\downarrow$ & $\delta_1$$\uparrow$ & AbsRel$\downarrow$ & $\delta_1$$\uparrow$ & AbsRel$\downarrow$ & $\delta_1$$\uparrow$ & AbsRel$\downarrow$ & $\delta_1$$\uparrow$ \\
    
    \midrule
    \multirow{9}{*}{ \rotatebox{90}{\textit{Discriminative}}} 
    & DiverseDepth\cite{diversedepth} & 11.7 & 87.5 & 19.0 & 70.4 & 22.8 & 69.4 & 10.9 & 88.2 & - & - \\
    & MiDaS\cite{midas}  & 11.1 & 88.5 & 23.6 & 63.0 & 18.4 & 75.2 & 12.1 & 84.6 & - & - \\
    & LeReS\cite{leres}  & 9.0 & 91.6 & 14.9 & 78.4 & 17.1 & 77.7 & 9.1 & 91.7 & - & - \\
    & Omnidata\cite{eftekhar2021omnidata}  & 7.4 & 94.5 & 14.9 & 83.5 & 16.6 & 77.8 & 7.5 & 93.6 & - & - \\
    & HDN\cite{zhang2022hierarchical}  & 6.9 & 94.8 & 11.5 & 86.7 & 12.1 & 83.3 & 8.0 & 93.9 & - & - \\
    & DPT\cite{dpt}  & 9.8 & 90.3 & 10.0 & 90.1 & 7.8 & 94.6 & 8.2 & 93.4 & - & - \\
    & Depth Anything v2~\cite{yang2024depthv2} & 4.1 & 97.6 & 8.0 & 94.0 & 4.6 & 97.9 & 4.2 & 97.6 & 8.0 & 95.2 \\
    & Depth Pro~\cite{depthpro} & 4.0 & 97.8 & 6.8 & 95.5 & 5.8 & 97.0 & 3.9 & 97.8 & 6.1 & 95.9 \\
    & MoGe 2~\cite{moge2} & \textbf{3.1} & \textbf{98.4} & \textbf{4.9} & \textbf{97.2} & \textbf{3.2} & \textbf{98.9} & \textbf{3.8} & \textbf{97.1} & \textbf{4.8} & \textbf{97.1} \\
    \midrule
    \multirow{6}{*}{ \rotatebox{90}{\textit{Generative}}} 
    & Marigold\cite{ke2024marigold}  & 5.5 & 96.4 & 9.9 & 91.6 & 6.5 & 96.0 & 6.4 & 95.1 & 10.0 & 90.7 \\
    & GeoWizard\cite{fu2025geowizard}  & 5.2 & 96.6 & 9.7 & 92.1 & 6.4 & 96.1 & 6.1 & 95.3 & 12.0 & 89.8 \\
    & DepthFM\cite{depthfm}  & 5.5 & 96.3 & 8.9 & 91.3 & 5.8 & 96.2 & 6.3 & 95.4 & - & - \\
    & GenPercept\cite{GenPercept}  & 5.2 & 96.6 & 9.4 & 92.3 & 6.6 & 95.7 & 5.6 & 96.5 & - & - \\
    & Lotus\cite{he2024lotus}  & 5.4 & 96.8 & 8.5 & 92.2 & 5.9 & 97.0 & 5.9 & 95.7 & 9.8 & 92.4 \\
    & PPD (Ours) & \textbf{3.3} & \textbf{98.2} & \textbf{5.3} & \textbf{97.0} & \textbf{3.0} & \textbf{99.1} & \textbf{3.5} & \textbf{98.1} & \textbf{5.2} & \textbf{97.0} \\
    \bottomrule
    \end{tabular}
    \label{tab:zeroshot_mde}
\end{table*}

\begin{table*}
    \caption{\textbf{Zero-shot video depth estimation.} Our PPVD achieves the best accuracy among all methods on four benchmarks. Unlike monocular depth estimation, video depth estimation requires aligning the predicted depth maps to the ground truth using a unified scale and shift across the entire video.}
    \small
    \centering
    \begin{tabular}{lcccccccc}
    \toprule
    \multirow{2}{*}{Method}  & \multicolumn{2}{c}{NYUv2} & \multicolumn{2}{c}{Scannet} & \multicolumn{2}{c}{Bonn} & \multicolumn{2}{c}{KITTI} \\

    \cmidrule(lr){2-3}\cmidrule(lr){4-5}\cmidrule(lr){6-7}\cmidrule(lr){8-9}
    
    ~ & AbsRel$\downarrow$ & $\delta_1$$\uparrow$ & AbsRel$\downarrow$ & $\delta_1$$\uparrow$ & AbsRel$\downarrow$ & $\delta_1$$\uparrow$ & AbsRel$\downarrow$ & $\delta_1$$\uparrow$ \\
    
    \midrule
    Depth Anything v2~\cite{yang2024depthv2} & 9.4 & 92.8 & 15.0 & 76.8 & 12.7 & 86.4 & 13.7 & 81.5 \\
    NVDS~\cite{nvds} & 21.7 & 59.8 & 20.7 & 62.8 & 19.9  & 67.4 & 23.3 & 61.4 \\
    ChoronDepth~\cite{chorondepth} & 17.3 & 77.1 & 19.9 & 66.5 & 19.9 & 66.5 & 24.3 & 57.6  \\
    DepthCrafter~\cite{depthcrafter} & 14.1 & 82.2 & 16.9 & 73.0 & 15.3 & 80.3 & 16.4 &75.3  \\
    RollingDepth~\cite{rollingdepth} & 8.9 & 92.4 & 10.2 & 90.1 & 8.8 & 93.1 & 10.7 & 88.7 \\
    Video Depth Anything~\cite{vda} & 6.2 & 97.1 & 8.9 & 92.6 & 7.1 & 95.9 & 8.3 & 94.4 \\
    PPVD (Ours) & \textbf{3.8} & \textbf{99.0} & \textbf{3.7} & \textbf{98.8} & \textbf{4.8} & \textbf{97.9} & \textbf{5.9} & \textbf{97.0} \\
    \bottomrule
    \end{tabular}
    \label{tab:zeroshot_vde}
\end{table*}

\section{Experiments}
\label{experiments}

\subsection{Experimental Setup}
\textbf{Training datasets.} Our objective is to estimate pixel-perfect depth maps, which, when converted to point clouds, are free of \textit{flying pixels} and geometric artifacts. To achieve this, it is essential to train on datasets with high-quality ground truth point clouds. Therefore, we mainly adopt Hypersim~\cite{hypersim}, because it is a photorealistic synthetic dataset with accurate and clean 3D geometry, which contains approximately 54K samples. We also additionally leverage four datasets, UrbanSyn~\cite{urbansyn} (7.5K), UnrealStereo4K~\cite{unrealstereo4k} (8K), VKITTI~\cite{vkitti} (25K), and TartanAir~\cite{tartanair} (30K), to further enhance the model’s generalization and robustness. For the video depth estimation, we further incorporate IRS~\cite{irs} (102K) and PointOdyssey~\cite{pointodyssey} (237K) to improve temporal consistency and motion robustness.

\textbf{Evaluation setup.} For monocular depth estimation, we align the predicted depth map with the ground truth by applying a scale and shift for each frame, and then evaluate the zero-shot monocular depth estimation performance on five real-world datasets: NYUv2~\cite{nyuv2}, KITTI~\cite{kitti}, ETH3D~\cite{eth3d}, ScanNet~\cite{scannet}, and DIODE~\cite{diode}, covering both indoor and outdoor scenes. For video depth estimation, we align the predicted depth maps with the ground truth by applying a unified scale and shift for the entire video, and then evaluate the zero-shot video depth estimation performance on four real-world video datasets: NYUv2~\cite{nyuv2}, ScanNet~\cite{scannet}, Bonn~\cite{bonn}, and KITTI~\cite{kitti}, with each scene containing 500 video frames. 

To evaluate the accuracy of depth estimation, we adopt two widely-used evaluation metrics: Absolute Relative Error (AbsRel) and $\delta_1$ accuracy. To demonstrate that our model predicts point clouds without \textit{flying pixels}, we convert the estimated depth maps into 3D point clouds and evaluate them using the proposed edge-aware metric. For monocular depth estimation, the ablation experiments are conducted using a $512 \times 512$ resolution models for simplicity, whereas the final models are fine-tuned at a resolution of $1024 \times 768$, achieving the best performance.

\subsection{Zero-Shot Monocular Depth Estimation}
To evaluate our monocular depth model PPD’s zero-shot generalization, we compare it with recent depth estimation models~\cite{yang2024depthv2,depthpro,ke2024marigold,he2024lotus,depthfm} on five real-world benchmarks. As shown in Table~\ref{tab:zeroshot_mde}, our PPD significantly outperforms all other generative depth estimation models for all evaluation metrics. Unlike previous generative models, we do not rely on image priors from a pretrained Stable Diffusion~\cite{stablediff} model. Instead, our diffusion model is trained from scratch and still achieves superior performance. Our PPD generalizes well to a wide range of real-world scenes, even when trained solely on synthetic depth datasets. Visual comparisons are shown in Figure~\ref{fig:demo}, our PPD preserves more fine-grained details than Depth Anything v2~\cite{yang2024depthv2} and MoGe 2~\cite{moge2}. Moreover, it demonstrates significantly higher robustness than Depth Pro~\cite{depthpro}, especially in challenging regions with complex textures, cluttered backgrounds, or large sky areas. Unlike previous models that use convolutional architectures, \eg, denoising U-Net for generative models and DPT for discriminative models, our model is purely transformer-based, with no convolutional layers.

\subsection{Zero-Shot Video Depth Estimation}
To evaluate the performance of our video depth model PPVD, we compare it with recent video depth estimation models~\cite{depthcrafter,depthanyvideo,rollingdepth,vda} on four real-world video benchmarks. As shown in Table~\ref{tab:zeroshot_vde}, our PPVD significantly outperforms previous generative and discriminative models, surpassing the previously best generative model RollingDepth~\cite{rollingdepth} by 63.7\% on ScanNet, and exceeding the previously best discriminative model Video Depth Anything~\cite{vda} by 58.4\%. Previous video depth estimation methods either impose temporal consistency constraints or leverage video priors from pretrained Stable Video Diffusion models. While these approaches can achieve visually consistent depth, their spatial accuracy remains limited. In contrast, the core of PPVD is to transform 3D geometry consistency into temporal consistency. Its semantic tokens encode both spatial relationship changes and camera poses, leading to a substantial improvement in depth estimation accuracy. Visual comparisons are shown in Figure~\ref{fig:pc_comp_video}. Our PPVD, while maintaining temporal consistency, produces significantly fewer \textit{flying pixels}.

\begin{table*}
    \caption{\textbf{Ablation studies for Pixel-Perfect Depth (PPD).} Inference time was tested on an RTX 4090 GPU.}
    \small
    \captionsetup{font=small}
    \centering
    \setlength\tabcolsep{0.9mm}
    \begin{tabular}{lccccccccccc}
    \toprule
    \multirow{2}{*}{Model}  & \multicolumn{2}{c}{NYUv2} & \multicolumn{2}{c}{KITTI} & \multicolumn{2}{c}{ETH3D} & \multicolumn{2}{c}{ScanNet} & \multicolumn{2}{c}{DIODE} & \multirow{2}{*}{Time(s)} \\

    \cmidrule(lr){2-3}\cmidrule(lr){4-5}\cmidrule(lr){6-7}\cmidrule(lr){8-9}\cmidrule(lr){10-11}
    
    ~ & AbsRel$\downarrow$ & $\delta_1$$\uparrow$ & AbsRel$\downarrow$ & $\delta_1$$\uparrow$ & AbsRel$\downarrow$ & $\delta_1$$\uparrow$ & AbsRel$\downarrow$ & $\delta_1$$\uparrow$ & AbsRel$\downarrow$ & $\delta_1$$\uparrow$ \\
    
    \midrule
    DiT (vanilla) & 22.5 & 72.8 & 27.3 & 63.9 & 12.1  & 87.4 & 25.7 & 65.1 & 23.9 & 76.5 & 0.19 \\
    DiT + REPA~\cite{repa} & 17.6 & 78.0 & 23.4 & 70.6 & 9.1 & 91.2 & 20.1 & 74.3 & 14.6 & 86.9 & 0.19 \\
    \midrule
    SP-DiT & 4.8 & 96.7 & 8.6 & 92.2 & 4.6 & 97.5 & 6.2 & 94.8 & 8.2 & 94.1 & 0.20  \\
    SP-DiT + Cas-DiT & \textbf{4.3} & \textbf{97.4} & \textbf{8.0} & \textbf{93.1} & \textbf{4.5} & \textbf{97.7} & \textbf{4.5} & \textbf{97.3} & \textbf{7.0} & \textbf{95.5} & \textbf{0.14}  \\

    \bottomrule
    \end{tabular}
    \label{tab:ablation_mde}
\end{table*}

\begin{table*}
    \caption{\textbf{Ablation studies on Vision Foundation Models (VFMs).} Note that we only utilize a pretrained encoder from these VFMs, such as a 24-layer ViT from DINOv2 or Depth Anything v2 (DAv2).}
    \small
    \centering
    \begin{tabular}{lcccccccccc}
    \toprule
    \multirow{2}{*}{VFM Type}  & \multicolumn{2}{c}{NYUv2} & \multicolumn{2}{c}{KITTI} & \multicolumn{2}{c}{ETH3D} & \multicolumn{2}{c}{ScanNet} & \multicolumn{2}{c}{DIODE} \\

    \cmidrule(lr){2-3}\cmidrule(lr){4-5}\cmidrule(lr){6-7}\cmidrule(lr){8-9}\cmidrule(lr){10-11}
    
    ~ & AbsRel$\downarrow$ & $\delta_1$$\uparrow$ & AbsRel$\downarrow$ & $\delta_1$$\uparrow$ & AbsRel$\downarrow$ & $\delta_1$$\uparrow$ & AbsRel$\downarrow$ & $\delta_1$$\uparrow$ & AbsRel$\downarrow$ & $\delta_1$$\uparrow$ \\
    
    \midrule
    DiT (vanilla) & 22.5 & 72.8 & 27.3 & 63.9 & 12.1  & 87.4 & 25.7 & 65.1 & 23.9 & 76.5 \\
    SP-DiT (MAE~\cite{he2022masked}) & 6.4 & 95.0 & 14.4  & 84.9 & 7.3  & 94.8 & 7.7 & 92.5 & 11.6 & 91.3 \\
    SP-DiT (DINOv2~\cite{dinov2}) & 4.8  & 96.4 & 9.3 & 91.2 & 5.6 & 96.2 & 5.1 & 96.9 & 9.2 &  93.5 \\
    SP-DiT (DAv2~\cite{yang2024depthv2}) & 4.3 & 97.4 & 8.0 & 93.1 & 4.5 & 97.7 & 4.5 & 97.3 & 7.0 & 95.5 \\
    SP-DiT (MoGe2~\cite{moge2}) & \textbf{3.3} & \textbf{98.2} & \textbf{5.3} & \textbf{97.0} & \textbf{3.0} & \textbf{99.1} & \textbf{3.5} & \textbf{98.1} & \textbf{5.2} & \textbf{97.0} \\
    \bottomrule
    \end{tabular}
    \label{tab:ablation_vfm}
\end{table*}

\subsection{Ablations and Analysis} 
\textbf{Component-wise ablation of PPD.} We adopt the vanilla DiT~\cite{dit} model as our baseline and conduct ablations on our proposed modules. Quantitative results are shown in Table~\ref{tab:ablation_mde}. Directly performing diffusion generation in high-resolution pixel space is highly challenging due to substantial computational costs and optimization difficulties, leading to significant performance degradation. As illustrated in Figure~\ref{fig:sp_dit}, the baseline model struggles with preserving global semantics and generating fine-grained visual details. To improve both training efficiency and performance, we utilize REPA~\cite{repa} to align intermediate tokens in DiT with a pretrained vision encoder~\cite{yang2024depthv2}. However, the resulting improvement remains very limited and still falls short of enabling pixel-space diffusion models to achieve performance comparable to state-of-the-art depth foundation models, such as Depth Anything v2~\cite{yang2024depthv2}. In contrast, the proposed Semantics-Prompted DiT (SP-DiT) addresses these challenges, achieving significantly improved accuracy, for example, a 78\% gain on the NYUv2 AbsRel metric. We further introduce a novel Cascaded DiT architecture (Cas-DiT) that progressively increases the number of tokens. This coarse-to-fine design not only significantly improves efficiency, for example, reducing inference time by 30\% on an RTX 4090 GPU, but also better models global context, leading to noticeable gains in accuracy.

\textbf{Ablations on vision foundation models (VFMs).} We evaluate the performance of SP-DiT using pretrained vision encoders from different VFMs, including MAE~\cite{he2022masked}, DINOv2~\cite{dinov2}, Depth Anything v2~\cite{yang2024depthv2}, and MoGe 2~\cite{moge2}, as illustrated in Table~\ref{tab:ablation_vfm}. All of them significantly boost performance.

\textbf{Component-wise ablation of PPVD.} Table~\ref{tab:ablation_vde} presents the component-wise ablation results of our PPVD. To extend PPD to long videos with minimal computational cost, we do not rely on the computationally expensive full attention over all input frames ($T \times H \times W$). Instead, we introduce a reference-guided token propagation (RGTP) strategy, as shown in Figure~\ref{fig:ppvd_network}. This strategy first assigns sparse (compressed) reference-frame tokens to all input frames, and then performs transformer operations on the single-frame tokens, i.e., $H \times W + (H/\pi) \times (W/\pi)$. Through these sparse reference tokens, we propagate the scene’s scale and shift information to all input frames. In our experiments, $\pi$ is set to 4. From the quantitative results in Table~\ref{tab:ablation_vde}, it can be seen that our RGTP significantly improves accuracy. Subsequently, we replace the single-view SP-DiT with the multi-view SC-DiT. SC-DiT provides view-consistent semantics, which also implicitly encodes camera poses, further enhancing depth estimation accuracy.

\begin{table*}
    \caption{\textbf{Ablation studies for Pixel-Perfect Video Depth (PPVD).} RGTP denotes the proposed Reference-Guided Token Propagation strategy.}
    \small
    \centering
    \begin{tabular}{lcccccccc}
    \toprule
    \multirow{2}{*}{Model}  & \multicolumn{2}{c}{NYUv2} & \multicolumn{2}{c}{ScanNet}& \multicolumn{2}{c}{Bonn} & \multicolumn{2}{c}{KITTI}\\

    \cmidrule(lr){2-3}\cmidrule(lr){4-5}\cmidrule(lr){6-7}\cmidrule(lr){8-9}
    
    ~ & AbsRel$\downarrow$ & $\delta_1$$\uparrow$ & AbsRel$\downarrow$ & $\delta_1$$\uparrow$ & AbsRel$\downarrow$ & $\delta_1$$\uparrow$ & AbsRel$\downarrow$ & $\delta_1$$\uparrow$ \\
    
    \midrule
    SP-DiT (DAv2~\cite{yang2024depthv2}) & 12.2 & 85.0 & 13.9 & 81.0 & 12.5 & 86.6 & 11.3 & 88.7 \\
    SP-DiT (DAv2~\cite{yang2024depthv2}) + RGTP & 7.6 & 95.2 & 8.8 & 93.2 & 7.9 & 96.0 & 8.6  & 93.7 \\
    SC-DiT (VGGT~\cite{vggt}) + RGTP & 4.5 & 98.6 & 5.3 & 97.9 & 5.3 & 97.8 & 6.9  & 95.8 \\
    SC-DiT ($\pi^3$~\cite{pi3}) + RGTP & \textbf{3.8} & \textbf{99.0} & \textbf{3.7} & \textbf{98.8} & \textbf{4.8} & \textbf{97.9} & \textbf{5.9} & \textbf{97.0} \\
    \bottomrule
    \end{tabular}
    \label{tab:ablation_vde}
\end{table*}

\subsection{Edge-Aware Point Cloud Evaluation}
Our objective is to estimate pixel-perfect depth maps that yield clean and accurate point clouds without \textit{flying pixels}, which often occur at object edges due to inaccurate depth predictions in these regions. However, existing evaluation benchmarks and metrics often struggle to reflect \textit{flying pixels} at object edges. For example, benchmarks like NYUv2 or KITTI usually lack edge annotations, while metrics such as AbsRel and $\delta_1$ are dominated by flat regions, making it difficult to assess depth accuracy at edges.

To address these limitations, we evaluate on the official test split of the Hypersim~\cite{hypersim} dataset, which provides high-quality ground-truth point clouds and is not used during training. We further propose an edge-aware point cloud metric that quantifies depth accuracy at edges. Specifically, we extract edge masks from ground-truth depth maps using the Canny operator and compute the Chamfer Distance between predicted and ground-truth point clouds near these edges.

Quantitative results in Table~\ref{tab:pc_eval} show that our PPD achieves the best performance. Since Hypersim does not provide video data, we restrict our evaluation to monocular depth estimation models only. Discriminative models like Depth Pro~\cite{depthpro} and Depth Anything v2~\cite{yang2024depthv2} tend to smooth edges, causing \textit{flying pixels}. Generative models such as Marigold~\cite{ke2024marigold} rely on VAE compression, which blurs edges and details, causing artifacts in the reconstructed point clouds. To illustrate this, we encode and decode the ground-truth depth using a VAE (GT(VAE)), without any generative process. Table~\ref{tab:pc_eval} and Figure~\ref{fig:vae} show that VAE compression introduces \textit{flying pixels}, leading to a larger Chamfer Distance than ours.

\begin{table*}
    \caption{\textbf{Edge-aware point cloud evaluation.} Our PPD achieves the best performance on the high-quality Hypersim test set.  To further verify that VAE compression leads to \textit{flying pixels}, we evaluate the ground truth depth maps after VAE reconstruction, denoted as GT(VAE).}
    \small
    \centering
    \begin{tabular}{cccccccc}
    \toprule
     & Marigold\cite{ke2024marigold} & GeoWizard\cite{fu2025geowizard}  & DepthAny. v2\cite{yang2024depthv2} & DepthPro\cite{depthpro} & MoGe 2\cite{moge2} & GT(VAE) & Ours\\
    \midrule
    Chamfer Distance $\downarrow$ & 0.17 & 0.16 & 0.18 & 0.14 & 0.13 & 0.12 & \textbf{0.07} \\
    \bottomrule
    \end{tabular}
    \label{tab:pc_eval}
\end{table*}

\section{Conclusion}
\label{conclusion}

We present Pixel-Perfect Visual Geometry Estimation models: \textbf{PPD} for monocular depth estimation and \textbf{PPVD} for video depth estimation. Both models utilize generative modeling in the pixel space to produce high-quality and flying-pixel-free point clouds from the estimated depth maps. Unlike previous generative depth estimation models, whether monocular or video-based, that rely on latent-space diffusion with a VAE, our models perform diffusion directly in the pixel space, thereby avoiding the \textit{flying pixels} caused by VAE compression. 

To overcome the high-dimensional optimization and training efficiency challenges inherent in pixel-space diffusion, and to further enhance accuracy and temporal consistency, we propose Semantics-Prompted DiT for PPD and Semantics-Consistent DiT for PPVD. These specialized DiT architectures significantly boost the accuracy and temporal consistency of our models. Additionally, a Cascaded DiT architecture is employed to further enhance their efficiency. Finally, our PPD and PPVD models achieve new state-of-the-art results among all generative monocular and video depth estimation models.

\bibliographystyle{IEEEtran}
\bibliography{ppvg}


\section{Biography Section}


\begin{IEEEbiography}[{\includegraphics[width=1in,height=1.25in,clip,keepaspectratio]{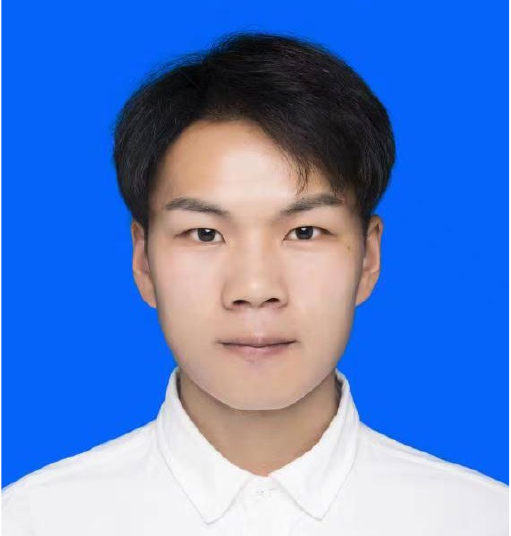}}]{Gangwei Xu}
is a PhD student at the Department of Electronic Information and Communications at Huazhong University of Science and Technology. He is supervised by Prof. Xin Yang. He received his B.Eng. degree from Huazhong University of Science and Technology in 2021. His current research focuses on depth estimation and 3D/4D reconstruction. He has published multiple papers in IEEE-TPAMI, NeurIPS, and CVPR. He also serves as a reviewer for top-tier journals and conferences, including IEEE-TPAMI, IJCV, NeurIPS, CVPR, etc.
\end{IEEEbiography}

\begin{IEEEbiography}
[{\includegraphics[width=1in,height=1.25in,clip,keepaspectratio]{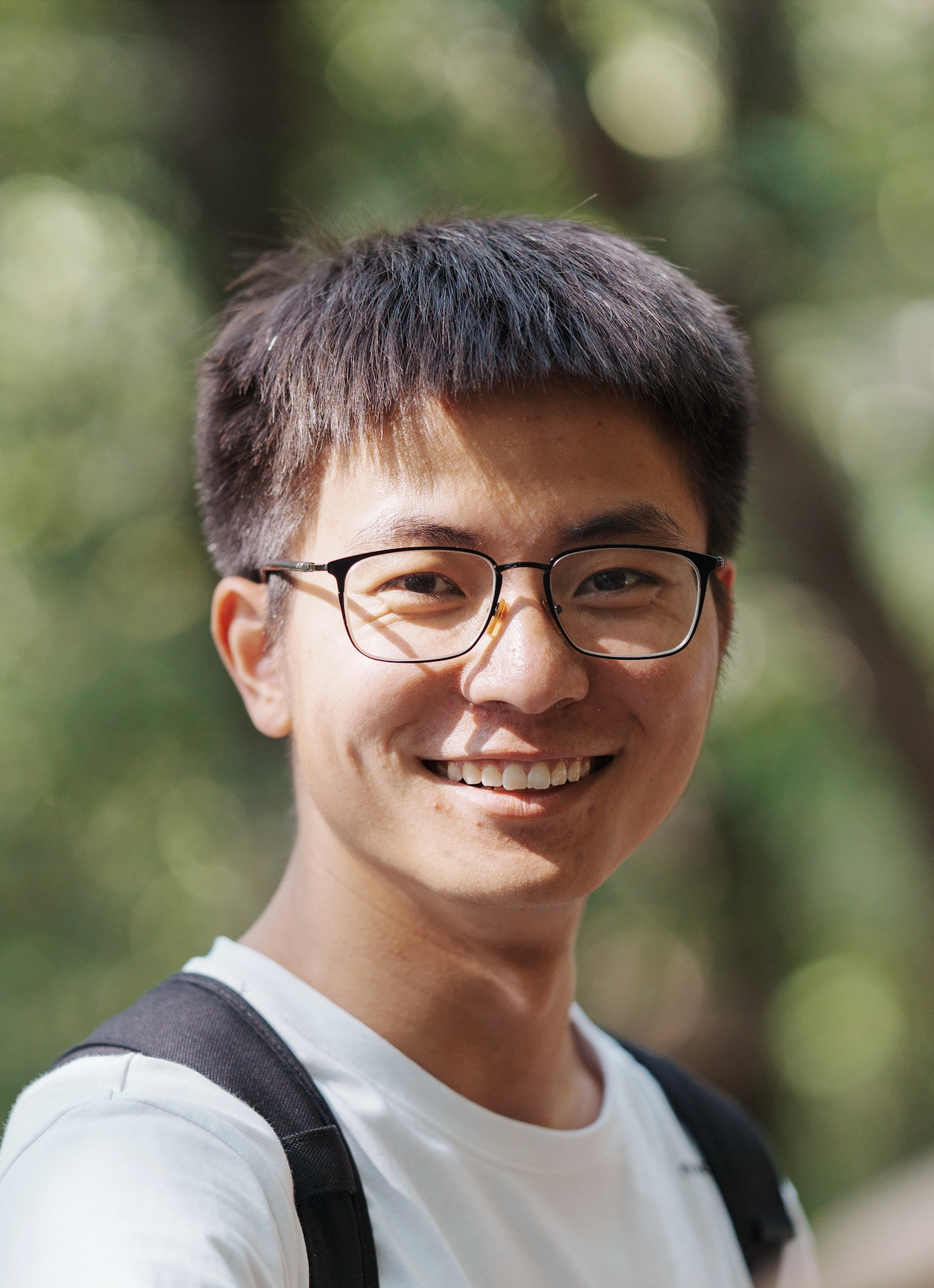}}]{Haotong Lin} is a PhD student in Computer Science at Zhejiang University, advised by Prof. Xiaowei Zhou. He obtained my bachelor degree in Computer Science from Zhejiang University in 2021. His current research focuses on depth estimation and 3D/4D reconstruction.
\end{IEEEbiography}

\begin{IEEEbiography}
[{\includegraphics[width=1in,height=1.25in,clip,keepaspectratio]{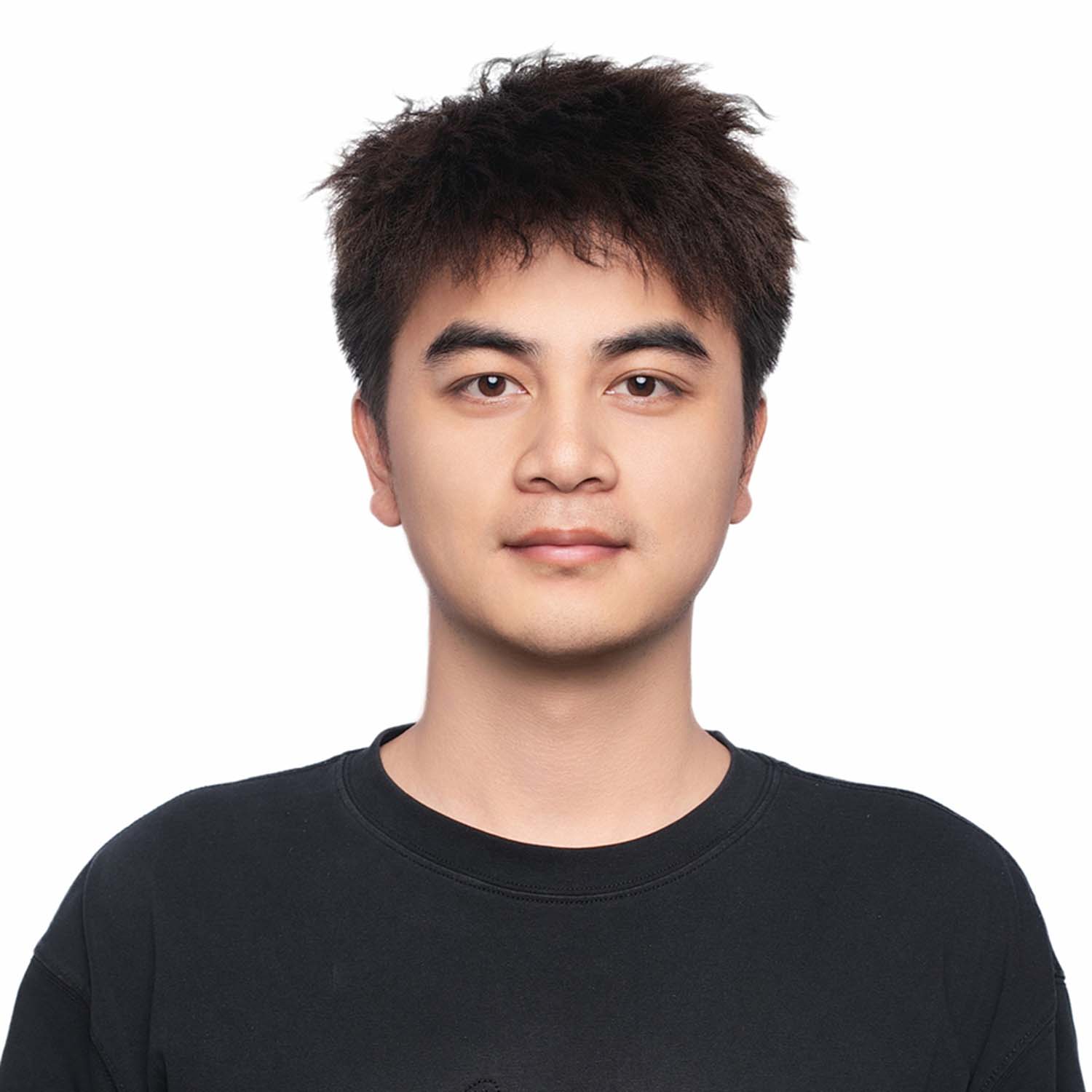}}]{Hongcheng Luo} received his master's degree from Huazhong University of Science and Technology in 2019. He is currently an Algorithm Researcher at Xiaomi EV. Prior to joining Xiaomi, he worked at Alibaba DAMO Academy.
\end{IEEEbiography}

\begin{IEEEbiography}
[{\includegraphics[width=1in,height=1.25in,clip,keepaspectratio]{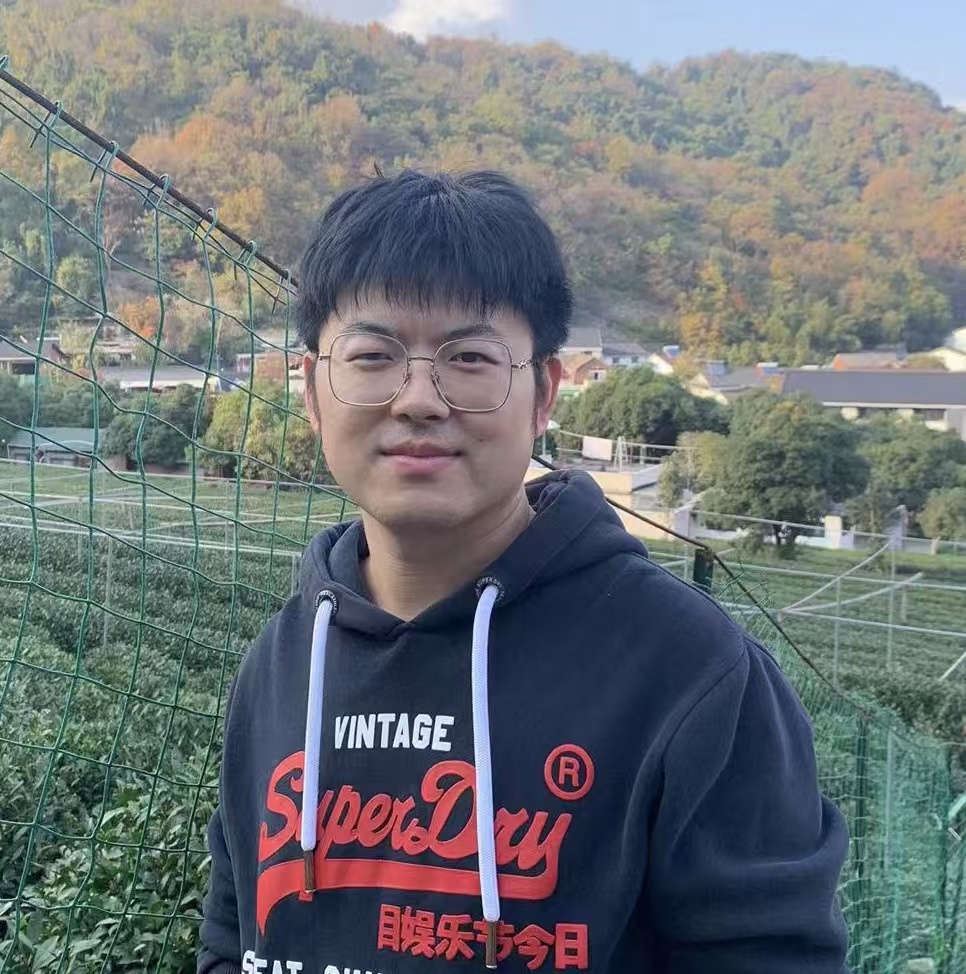}}]{Haiyang Sun} received the Master degree in information and communication engineering from Tsinghua University, Beijing, China, in 2016. He is currently an Expert Algorithm Engineer at XiaomiEV. His research interests include World Model, 3D vision and Autonomous Driving.
\end{IEEEbiography}

\begin{IEEEbiography}
[{\includegraphics[width=1in,height=1.25in,clip,keepaspectratio]{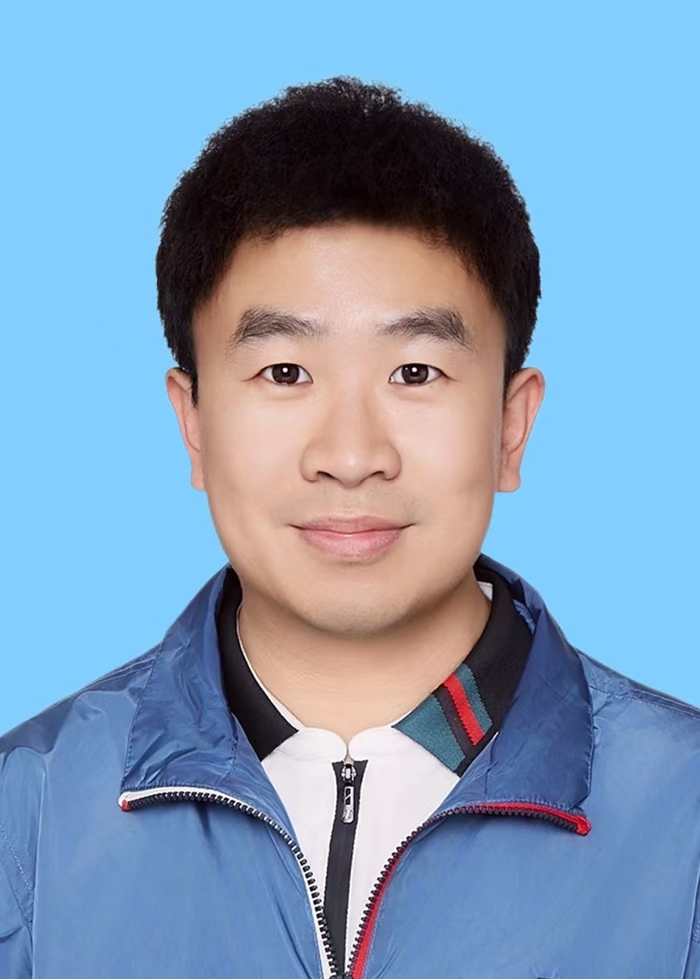}}]{Bing Wang} received his Ph.D. degree from School of Electrical and Electronic Engineering, Nanyang Technological University, Singapore, in 2016. He is currently an Expert Algorithm Engineer at Xiaomi EV. His research interests include computer vision, machine learning, world model, autonomous driving and robotics.
\end{IEEEbiography}

\begin{IEEEbiography}
[{\includegraphics[width=1in,height=1.25in,clip,keepaspectratio]{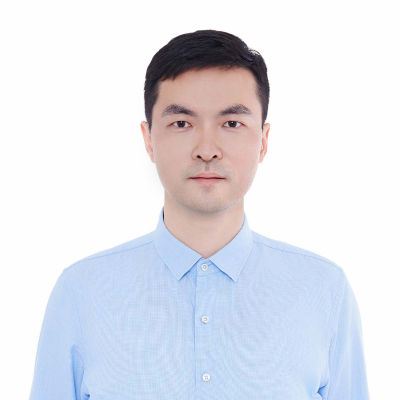}}]{Guang Chen} received the PhD degree from Electrical and Computer Department of the University of Missouri, in 2014. He is now an Expert Algorithm Engineer at Xiaomi EV. His research interests include computer vision,  machine learning and autonomous driving.
\end{IEEEbiography}

\begin{IEEEbiography}
[{\includegraphics[width=1in,height=1.25in,clip,keepaspectratio]{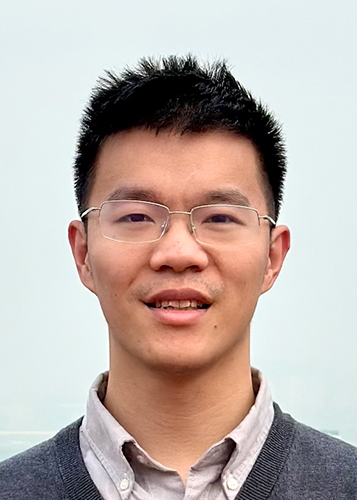}}]
{Sida Peng} received the PhD degree from the College of Computer Science and Technology, Zhejiang University, in 2023. He is a research professor with the School of Software, Zhejiang University, China. His research interests include volumetric video, driving simulator and egocentric intelligence.
\end{IEEEbiography}

\begin{IEEEbiography}
[{\includegraphics[width=1in,height=1.25in,clip,keepaspectratio]{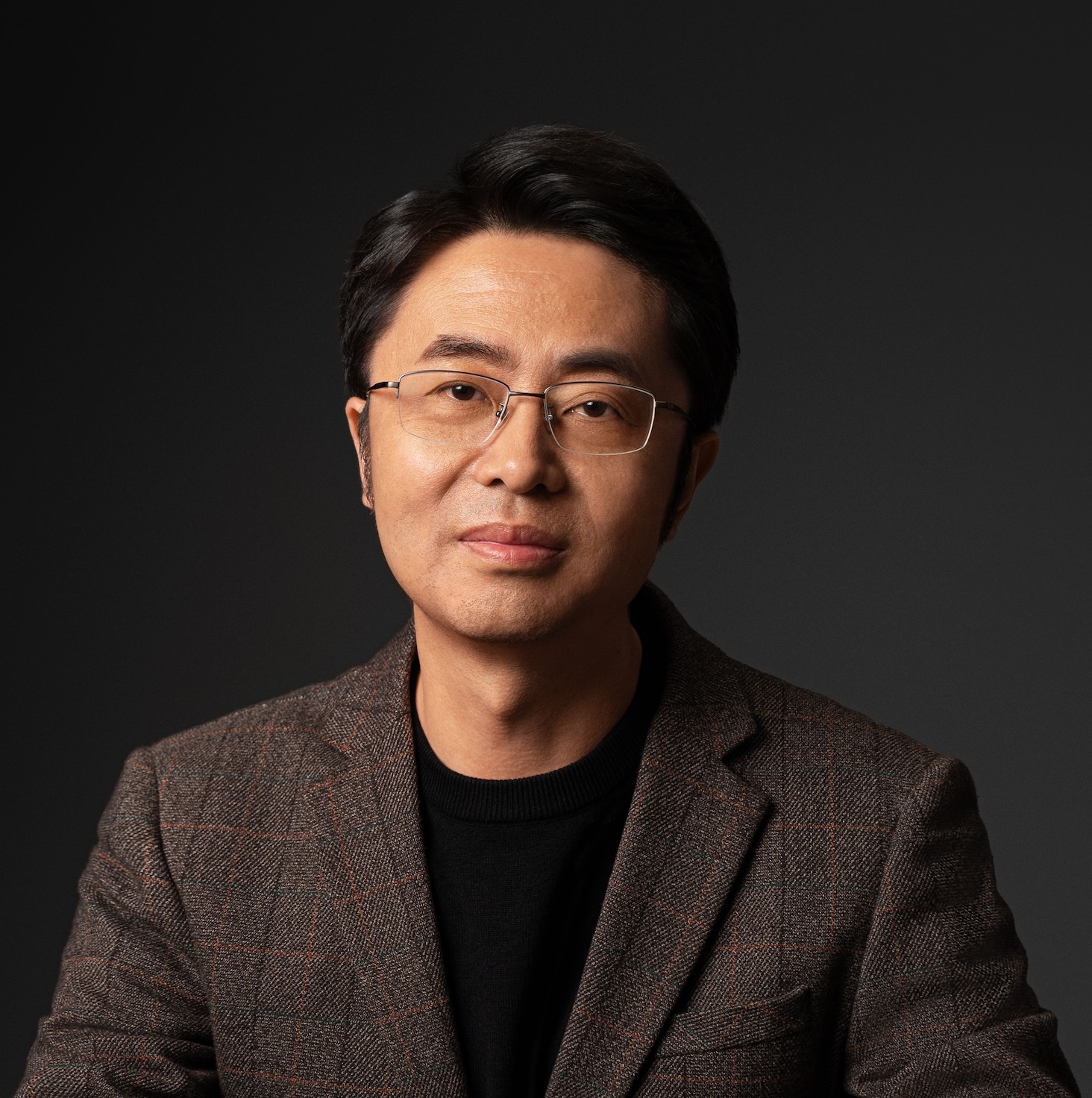}}]{Hangjun Ye} received his Ph.D. degree from the Department of Computer Science and Technology, Tsinghua University, China, in 2003. He is currently the head of the Autonomous Driving and Robotics Division, Xiaomi EV. His research interests include computer vision, machine learning, autonomous driving and robotics.
\end{IEEEbiography}

\begin{IEEEbiography}[{\includegraphics[width=1in,height=1.251in,clip,keepaspectratio]{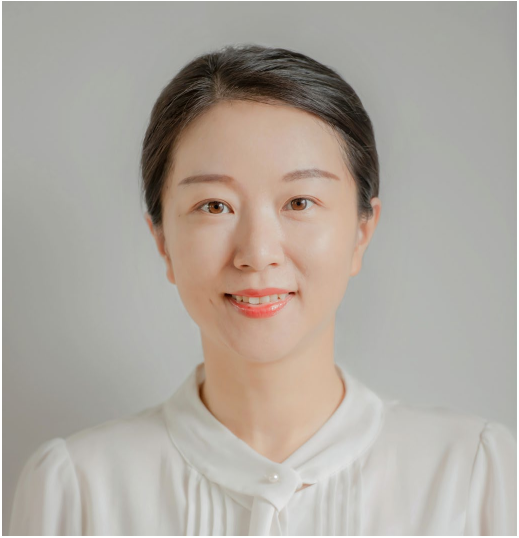}}]{Xin Yang}
is a Professor at the Department of Electronic Information and Communications at Huazhong University of Science and Technology. She received her Ph.D. degree in the Department of Electrical Computer Engineering at the University of California, Santa Barbara (UCSB).  Her research interests include medical image analysis and 3D vision. She is the recipient of the National Natural Science Fund of China for Excellent Youth Scholar and China Society of Image and Graphics Qingyun Shi Female Scientist Award. She has published over 90 technical papers and held 20 patents. She serves as an Associate Editor of IEEE-TVCG, IEEE-TMI and Multimedia System, an Area Chair of CVPR’24, MICCAI’19-21, and ACM MM’18. She is also a reviewer of top-tier journals such as IEEE-TPAMI, IJCV, etc.
\end{IEEEbiography}

\vfill

\end{document}